\documentclass[10pt,twocolumn,letterpaper]{article}

\usepackage[pagenumbers]{cvpr} 

\usepackage{xcolor}
\usepackage{amsfonts}  
\usepackage{multirow}
\usepackage{comment}
\usepackage{algpseudocode}
\usepackage{amsmath}
\usepackage{amssymb}
\usepackage[linesnumbered,ruled,vlined]{algorithm2e}

\definecolor{cvprblue}{rgb}{0.21,0.49,0.74}
\usepackage[pagebackref,breaklinks,colorlinks,allcolors=cvprblue]{hyperref}

\def\paperID{} 
\def\confName{CVPR}
\def\confYear{2026}

\title{Gaze-Regularized Vision-Language-Action Models for Robotic Manipulation}

\author{Anupam Pani\\
Institute of Data Science\\
University Of Hong Kong \\
\and
Yanchao Yang\\
Institute of Data Science\\
University Of Hong Kong \\
}

\begin{document}
\maketitle
\begin{abstract}

Despite advances in Vision-Language-Action (VLA) models, robotic manipulation struggles with fine-grained tasks because current models lack mechanisms for active visual attention allocation. Human gaze naturally encodes intent, planning, and execution patterns -- offering a powerful supervisory signal for guiding robot perception. We introduce a gaze-regularized training framework that aligns VLA models' internal attention with human visual patterns without architectural modifications or inference-time overhead. Our method transforms temporally aggregated gaze heatmaps into patch-level distributions and regularizes the transformer's attention through KL divergence, creating an inductive bias toward task-relevant features while preserving deployment efficiency. When integrated into existing VLA architectures, our approach yields 4-12\% improvements across manipulation benchmarks.
The gaze-regularized models reach equivalent performance with  fewer training steps and maintain robustness under lighting variations and sensor noise.
Beyond performance metrics, the learned attention patterns produce interpretable visualizations that mirror human strategies, enhancing trust in robotic systems. Moreover, our framework requires no eye-tracking equipment and applies directly to existing datasets. These results demonstrate that human perceptual priors can significantly accelerate robot learning while improving both task performance and system interpretability.

\end{abstract}

\section{Introduction}
\label{sec:introduction}

Vision-Language-Action (VLA) models 
have emerged as a powerful paradigm 
for robotic manipulation, 
leveraging large-scale pretraining 
to enable natural language-conditioned 
control of complex behaviors~\cite{liu2024robomambaefficientvisionlanguageactionmodel,embodimentcollaboration2025openxembodimentroboticlearning,ahn2022icanisay,brohan2023rt2visionlanguageactionmodelstransfer,black2024pi0visionlanguageactionflowmodel,kim2024openvlaopensourcevisionlanguageactionmodel}. 
By combining visual perception 
with linguistic understanding, 
these models translate high-level instructions 
into precise robot actions, 
offering unprecedented flexibility for deployment in assistive robotics and human-machine collaboration~\cite{li2024visionlanguagefoundationmodelseffective,zhao2024vialmsurveybenchmarkvisually}. {\it However,} despite their architectural sophistication and vast pretraining data, 
current VLA approaches face fundamental challenges 
that limit their practical deployment in unstructured environments.



The core limitation 
of existing VLA models 
lies in their passive visual understanding. 
While robust perception 
for embodied agents requires actively seeking 
task-relevant information, 
current models tend to 
process all visual information as an entirety, 
attempting to simultaneously learn {\it where} 
to look and {\it how} to act. 
This joint learning burden 
results in inefficient training, 
slow convergence, and suboptimal performance even with millions of demonstrations~\cite{Wang_2025,poria202510openchallengessteering}. 
Despite powerful vision-language backbones 
from pretraining, 
these models discover relevant visual regions entirely through trial-and-error, 
lacking the selective attention mechanisms crucial for efficient manipulation.


\begin{figure}[!t]
    \centering
    \includegraphics[width=1.0\linewidth]{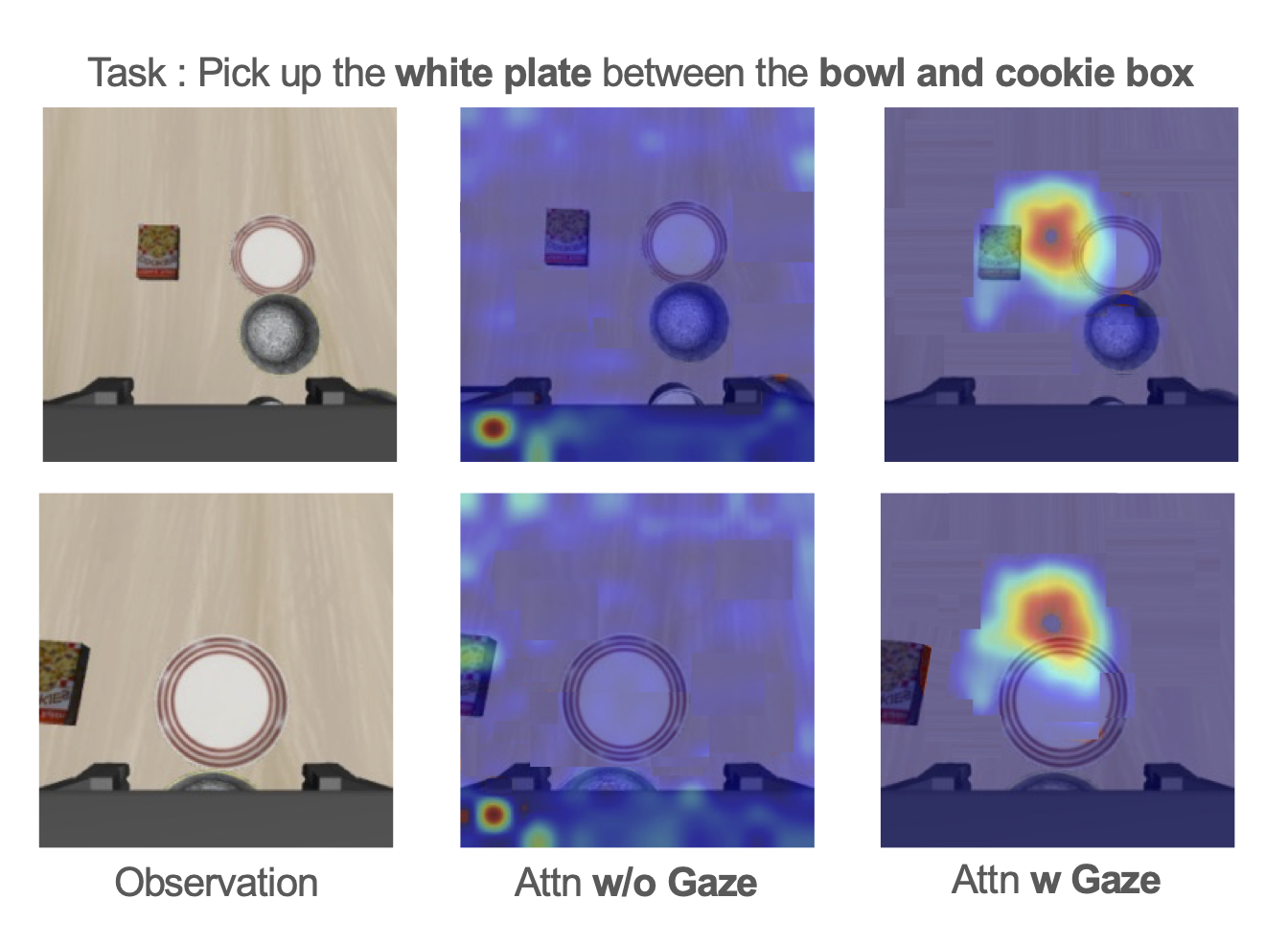}
    \vspace{-5mm}
    \caption{
    \textbf{Effect of Gaze Regularization.} 
    The baseline (middle)
    exhibits scattered attention 
    across the scene, 
    while the gaze-regularized model (right) concentrates on task-relevant regions (the plate and its immediate surroundings). 
    This focused attention pattern 
    not only improves task performance 
    but also provides interpretable visual grounding that enhances trust in the model.
    }
    \vspace{-3mm}
    \label{fig:teaser}
\end{figure}

Consider an assistive robot 
retrieving a specific medicine bottle 
from a cluttered cabinet, 
or a manufacturing robot selecting a precise component from a bin of similar parts. 
In these scenarios, 
the inability to focus visual attention 
on task-critical features 
leads directly to failure. 
Our experiments reveal that baseline VLA models 
plateau at approximately 86\% success rate on spatial manipulation tasks, 
suggesting systematic limitations in identifying task-relevant information for further improvement. 
Moreover, this lack of interpretable attention mechanisms makes robots difficult to trust, 
creating a reliability gap that hinders adoption in safety-critical applications (Fig.~\ref{fig:teaser}).


Humans naturally solve this attention allocation problem via selective visual perception, 
rapidly identifying and tracking task-relevant regions while filtering distractions. 
Eye-tracking studies consistently show that human gaze during manipulation exhibits strong regularities, 
with fixations concentrating on manipulated objects, 
upcoming targets, and critical spatial boundaries~\cite{Frischen2007,Tipper2010}. 
These patterns encode a temporal sequence of scan, plan, and act that characterizes skilled manipulation, with fixations often preceding hand movements to reveal anticipatory intent -- information that passive visual processing cannot capture.


Therefore, 
we propose a gaze-regularized training strategy that leverages human visual attention patterns to transform VLA models from passive observers to active perceivers, 
without requiring architectural modifications or runtime dependencies. 
Our {\it key} insight is that human gaze provides rich supervisory signals encoding both perceptual relevance and action-oriented intent, which can shape the model's internal attention mechanisms toward task-relevant regions during training~\cite{Hayhoe2003-gy,Belardinelli2016-sx,Belardinelli2016-wh}. 
Correspondingly, 
our approach aligns the transformer's vision-language attention distributions with human fixation patterns through a training-only regularization framework.


Our method first addresses 
the practical challenge that robotic datasets rarely include human eye-tracking data by employing a pretrained gaze prediction model to generate synthetic gaze heatmaps. 
These heatmaps capture both instantaneous fixations and anticipatory gaze shifts that characterize human manipulation behavior, 
effectively encoding the scan-plan-act sequence underlying skilled task execution. 
We then convert these continuous heatmaps into discrete probability distributions over visual tokens, enabling direct regularization of the transformer's attention mechanism through Kullback-Leibler divergence minimization. 
The gaze regularization term augments the standard training objective, creating a soft inductive bias that guides attention while allowing learning 
task-specific patterns.


Our approach achieves substantial improvements across all benchmarks while maintaining inference-time efficiency. 
The gaze-regularized model reaches 95.5\% success on LIBERO-Spatial versus 85.9\% for baseline, 
with comparable gains on Object and Goal suites. 
These improvements emerge early --
with 6-8\% gains at just 20,000 training steps -- 
demonstrating superior sample efficiency, 
and persist across different environments and visual perturbations. 
Crucially, all benefits occur without inference modifications, preserving real-time deployment while providing interpretable attention maps that enhance human trust.


Our contributions are threefold. 
{\it First,} 
we identify and formalize 
the passive perception limitation 
in VLA models and demonstrate how human gaze patterns can transform them into active perceivers. {\it Second,} 
we develop a practical gaze regularization framework that 
operates entirely during training 
without requiring eye-tracking equipment or architectural modifications, 
making it immediately deployable to existing systems. 
{\it Third,} 
we provide comprehensive experimental validation showing that gaze regularization consistently improves task performance, 
accelerates convergence, and enhances robustness across diverse scenarios. 
These results establish human visual attention 
as a valuable supervisory signal 
for efficient policy learning 
and 
suggest incorporating human perceptual strategies into VLA models.

\section{Related Work}
\label{sec:related_work}

\vspace{-2mm}
\paragraph{Gaze for Task Segmentation and Structure}
Human gaze has been leveraged to infer hierarchical task structure in robotics, as well as action recognition and prediction ~\cite{luo2025mindeyeomniassistgazedrivenllmenhanced,pmlr-v210-shen23a,takizawa2025gazeguidedtaskdecompositionimitation,10.1109/humanoids.2015.7363576,Li_2018_ECCV,8305456}. Takizawa et al.~\cite{takizawa2025gazeguidedtaskdecompositionimitation} showed that gaze fixation transitions during teleoperation provide robust signals for segmenting demonstrations into sub-tasks, simplifying long-horizon policy learning. However, these approaches use gaze solely for offline temporal segmentation rather than modulating perceptual attention during action generation, limiting their impact on visual reasoning.

\vspace{-2mm}
\paragraph{Gaze-Informed Perception and VLMs}
Gaze has been integrated directly into perceptual models to align them with human visual priors~\cite{Huang_2020,zhou2024learningobservergazezeroshotattention,9423298,10.1007/978-3-031-35596-7_29,10.5898/jhri.6.1.admoni,10.1109/lra.2021.3059619,10.3389/fpsyg.2015.01049,8593580,sood-etal-2021-vqa}. Yan et al.~\cite{yan2023voilaaaligningvisionlanguagemodels} modulated transformer attention keys with gaze heatmaps to ground reasoning in human-attended regions, though requiring runtime gaze input limits practicality. In robotics, Li et al.~\cite{li2024virt} introduced ``robotic gaze'' through dynamic zooming, while others explored other techniques like image cropping and foveated imagery for task-relevant focus~\cite{9103290,Kim_2021,11150769,chuang2025lookfocusactefficient}. These methods require architectural modifications or mappings, whereas our approach uses gaze as training-only supervision without altering model architecture or requiring inference-time gaze.

\vspace{-2mm}
\paragraph{Gaze as Supervisory Signal}
Training-time gaze supervision has shown promise in various domains~\cite{thammineni2020selectiveeyegazeaugmentationenhance,zhang2018agillearningattentionhuman,zhang2019atariheadatarihumaneyetracking,saran2021efficientlyguidingimitationlearning,li2025gazeguidedlearningavoidingshortcut,rong2021humanattentionfinegrainedclassification,alacam-etal-2024-eyes,chen2025gazeinsightbridginghuman,khaertdinova2024gazeassistedmedicalimagesegmentation}. Saran et al.~\cite{saran2021efficientlyguidingimitationlearning} introduced Coverage-based Gaze Loss for imitation learning, using gaze as weak supervision to guide attention in 2D Atari games while maintaining gaze-free inference. Similarly ~\cite{panigaze,mathew2025gazevlmvisionlanguagemodelmultitask} regularized attention in vision-language models using ground-truth gaze during training. 
We build on these concepts 
and enhance complex 6-DoF manipulation with temporal gaze aggregation and direct integration into VLA attention mechanisms.

\vspace{-2mm}
\paragraph{Multi-View Robotic Policies and Attention Mechanisms}
Our foundation stems from 
recent advances in scalable robotic policies. SAM2Act~\cite{fang2025sam2actintegratingvisualfoundation} leverages visual foundation models for efficient spatial representation in manipulation. VLA models like Pi-0, RT-2, OpenVLA, and RoboMamba etc.~\cite{black2024pi0visionlanguageactionflowmodel,kim2024openvlaopensourcevisionlanguageactionmodel,liu2024robomambaefficientvisionlanguageactionmodel,octomodelteam2024octoopensourcegeneralistrobot,brohan2023rt1roboticstransformerrealworld} demonstrate generalizable, language-conditioned control through internal attention mechanisms. 
Our contribution is modular with respect to these architectures -- we align their attention maps with human gaze patterns through auxiliary loss, enhancing perceptual grounding without structural changes,
{\it thus,} introducing a general framework for using temporally aggregated human gaze as a training-time regularizer for VLA models.

\section{Method}
\label{sec:method}

\begin{figure*}[!t]
  \centering
   \includegraphics[width=0.9\textwidth]{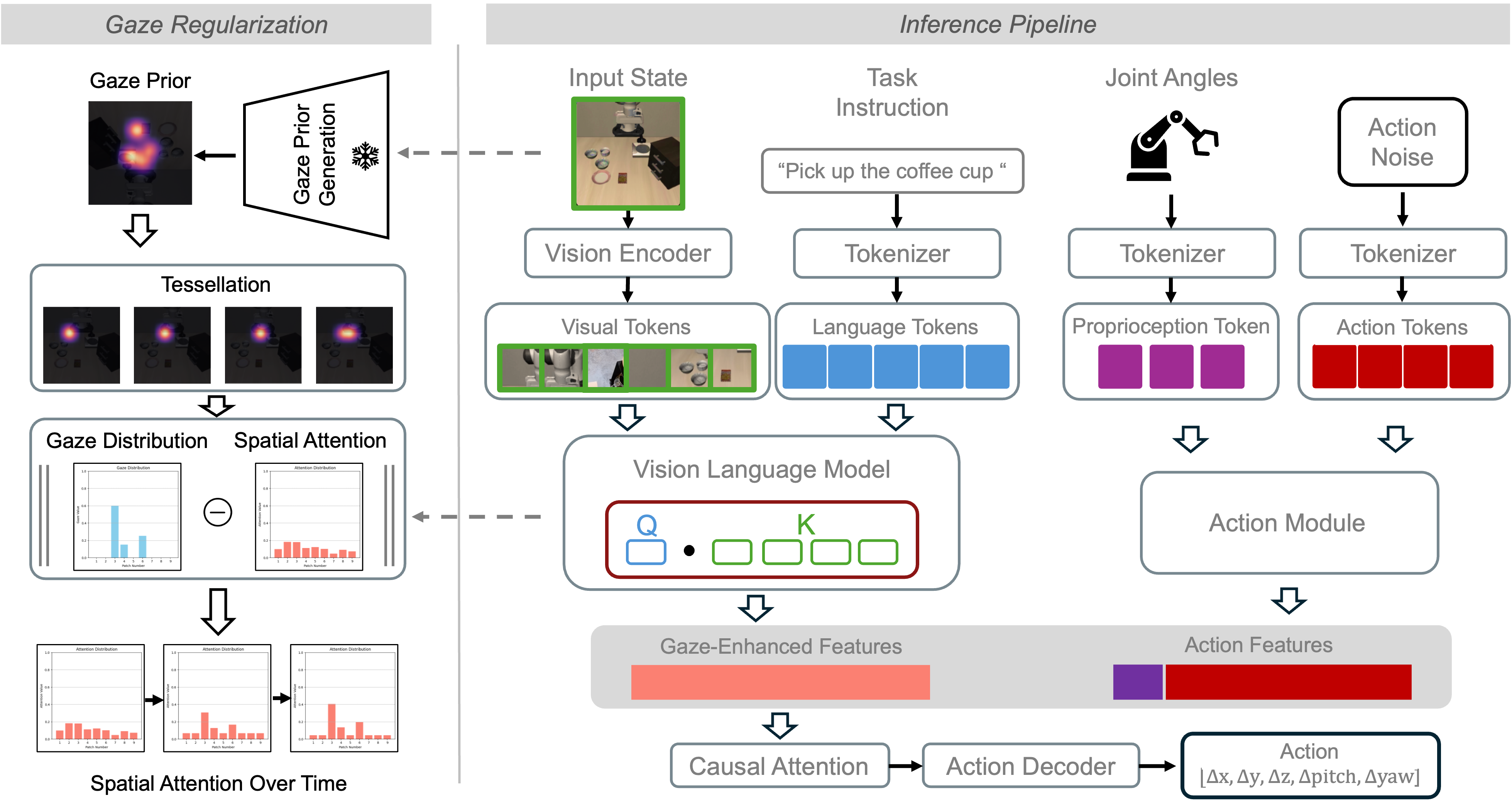}
\caption{
\textbf{Overview of the Proposed Gaze-Regularized VLA Framework.}
\textbf{Left:} During training, gaze priors are converted into patch-level gaze distributions that match the transformer’s attention resolution. 
The KL divergence between gaze and model attention is minimized, guiding the model to align its visual focus with human fixation patterns over time. 
\textbf{Right:} During inference, the policy operates without any gaze input. 
Visual, language, and proprioceptive tokens are processed by the vision–language backbone 
and action head, and fused through causal attention to produce action features, 
which are mapped by the action decoder 
to  control outputs. 
This training-time regularization yields gaze-aligned internal representations while maintaining a lightweight, gaze-free inference pipeline.
}
\vspace{-3mm}
   \label{fig:architecture}
\end{figure*}

To enhance 
the training efficiency and generalization 
of Vision-Language Action (VLA) models, 
we propose a gaze-regularized training strategy 
that utilizes human gaze 
as a training-time supervisory prior 
to direct the model’s internal attention 
to actionable regions in the scene, 
{\it without} changing the underlying architecture 
or requiring gaze at inference.
Next, 
we first formalize the standard 
VLA control problem 
and its attention structure (Sec.~\ref{sec:problem-formulation}). 
We then describe 
how we obtain human gaze priors for robotic data 
in the form of heatmaps, 
and convert them into distributions 
aligned with the transformer’s visual tokens (Sec.~\ref{sec:gaze_priors}). 
Further, 
we show how these gaze distributions 
are used to regularize the vision-language attention within the causal transformer backbone (Sec.~\ref{sec:attention-modulation}). 
Finally, 
we define the full training objective 
and inference-time procedure (Sec.~\ref{sec:training-objective}).

\subsection{Problem Formulation}
\label{sec:problem-formulation}

We treat the VLA policy 
as a neural network that predicts 
temporally extended robot actions 
conditioned on multi-modal observations.
Specifically,
let the policy parameters be $\theta$. 
At time step $t$, 
the model predicts a short-horizon action sequence:
\begin{equation}
A_t = [a_t, a_{t+1}, \ldots, a_{t+h-1}], \qquad h = 50,
\end{equation}
given multimodal input:
\begin{equation}
o_t = \{ I_t^{1:n}, \, \ell_t, \, q_t \},
\end{equation}
where $I_t^{1:n} = \{ I_t^1, \ldots, I_t^n \}$ represents the RGB frames 
from $n$ camera views, 
$\ell_t = [w_1, \ldots, w_T]$ denotes the tokenized language instruction, 
and $q_t \in \mathbb{R}^{d_p}$ encodes proprioceptive features such as joint angles and gripper pose.
The VLA policy 
is then trained to model the conditional distribution of future actions:
\begin{equation}
p_\theta(A_t \mid o_t).
\label{eq:policy_distribution}
\end{equation}
Each modality is first embedded 
by its respective encoder 
and projected into a shared latent space. 
The concatenated embeddings 
form the input token sequence 
to a transformer-based architecture $\pi_\theta$, 
which produces action predictions 
through causal attention across modalities.

Internally, 
the transformer backbone 
from the vision-language model 
produces a spatial attention distribution 
\( S_t^i \in \mathbb{R}^{N_v} \) for each view $i$ 
over visual tokens 
conditioned on a global representation 
of the language tokens. 
This distribution reflects 
how the language instruction attends 
to different visual patches, 
indicating relevance.


We hypothesize that 
aligning a VLA’s internal attention 
with human gaze distribution 
can improve both learning efficiency 
and downstream task performance 
in robotic manipulation.
Since eye gaze reflects 
how humans allocate visual attention 
to relevant regions 
before and during action execution, 
by shaping transformer attention towards these regions, 
the model shall acquire an inductive bias 
that mirrors human strategies 
for selective perception and control.
{\it Therefore,} our objective 
is to regularize 
this spatial attention using human gaze priors. 
During training, 
for each view $I_t^i$, 
a gaze prediction model produces a heatmap $H_t^i$, 
which is converted into a normalized, 
patch-level gaze distribution $G_t^i$ (Sec.~\ref{sec:gaze_priors}). 
The alignment between $S_t^i$ and $G_t^i$ 
forms the basis of our \textbf{gaze-regularization loss}, 
encouraging the model 
to allocate attention to regions 
humans naturally fixate on during manipulation. Next, we describe how the human gaze prior for robotic data is obtained.

\subsection{Gaze Prior Generation for Robotic Data}
\label{sec:gaze_priors}

A key challenge 
in leveraging gaze 
for VLA policy learning 
is the scarcity of robotic datasets 
that include human eye-tracking labels.
To address this, 
we augment standard robotic manipulation datasets 
with \emph{synthetic gaze} generated from gaze prediction models.
We denote this model as 
$\phi_{\text{gaze}}$, 
mapping visual frames 
to 
gaze heatmaps.

\vspace{-2mm}
\paragraph{Gaze Heatmap Generation.}
Given a short video clip 
$\{ I_{t-k}^i, \ldots, I_t^i \}$ from camera view~$i$, 
the pretrained gaze estimator 
produces a fixation heatmap 
that represents the likelihood of human visual attention at each pixel:
\begin{equation}
{[H_{t-k}^i,\ldots,H_{t}^i}] = \phi_{\text{gaze}}\big( \{ I_{t-k}^i, \ldots, I_t^i \} \big) \in \mathbb{R}^{k \times H_g \times W_g}.
\label{eq:gaze_heatmap}
\end{equation}
Each pixel intensity $H_t^i(x, y)$ encodes the predicted fixation probability at location $(x, y)$. 
To ensure quality, 
we employ the Global-Local Correlation (GLC) network~\cite{lai2022eye} 
due to its trustworthy performance on temporally contextual gaze prediction in egocentric video. 
Then we describe how the predicted heatmap 
is converted to the gaze distribution.

\vspace{-2mm}
\paragraph{From Heatmaps to Patch-Level Distributions.}
Since the VLA model operates 
on a fixed number of visual tokens 
rather than individual pixels,
we convert each gaze heatmap 
\(H_t^i\) into a 
\emph{patch-level probability distribution} \(G_t^i\)
that matches the spatial granularity 
of the attention map obtained from the transformer. 
Namely, we project each heatmap $H_t^i$ onto the same patch grid used by the vision encoder.
This process turns the raw pixel intensities of the gaze heatmap into a normalized distribution
indicating how likely each patch was within the human observer’s focus.

Let \(\mathbf{\Omega}\) denote 
the spatial domain of the gaze heatmap \(H_t^i\).
We divide \(\mathbf{\Omega}\) into a grid of \(N_v\) non–overlapping patches
\(\{\mathbf{p}_1, \mathbf{p}_2, \ldots, \mathbf{p}_{N_v}\}\)
corresponding to the same spatial partition used by the vision encoder
(e.g., a \(16{\times}16\) patch grid for \(N_v = 256\)).
Each patch \(\mathbf{p}_j\) represents the spatial region corresponding to a single visual token.

The gaze likelihood for each patch 
is computed by averaging the heatmap values within its region:
\begin{equation}
G_{t,j}^i = \frac{1}{Z} \sum_{(x, y) \in \mathbf{p}_j} H_t^i(x, y),
\qquad
Z = \sum_{x,y}H_t^i(x, y),
\label{eq:heatmap_to_token}
\end{equation}
where \(Z\) is a normalization constant ensuring the summation term
\(\sum_{j=1}^{N_v} G_{t,j}^i = 1\).

The resulting vector
\(G_t^i = [G_{t,1}^i, G_{t,2}^i, \ldots, G_{t,N_v}^i] \in \mathbb{R}^{N_v}\),
forms a discrete probability distribution over the transformer’s visual tokens,
making it directly comparable to the model’s internal attention map \(S_t^i\) for gaze–attention alignment. 
Next, we aggregate these heatmaps 
to capture human attention more comprehensively. 

\vspace{-2mm}
\paragraph{Temporal Aggregation of Gaze Heatmaps.}
Human gaze evolves over time 
and can shift frequently 
\emph{in anticipation} of upcoming actions.
Instead of a sub-sampling of frames 
that could miss critical gaze information during training, 
we aggregate the gaze heatmaps 
across a temporal window 
to obtain a more stable and comprehensive prior. 
Let $\{ H_{t-T}^i, \ldots, H_{t+T}^i \}$ denote the sequence of predicted gaze heatmaps 
within a temporal context of $2T{+}1$ frames centered at time~$t$. 
We compute a weighted temporal average:
\begin{equation}
\tilde{H}_t^i = \sum_{\delta=-T}^{T} w_{\delta} \, H_{t+\delta}^i,
\qquad
\sum_{\delta=-T}^{T} w_{\delta} = 1,
\label{eq:temporal_agg}
\end{equation}
where the coefficients $w_{\delta}$ assign highest importance to the current frame 
while incorporating adjacent frames to capture short-term anticipation.
The aggregated map $\tilde{H}_t^i$ therefore encodes both momentary fixations and anticipatory gaze shifts,
producing a smoother and temporally consistent signal than single-frame estimates.
Finally, $\tilde{H}_t^i$ is normalized using Eq.~\ref{eq:heatmap_to_token} 
to yield the aggregated token-level distribution $\tilde{G}_t^i$.
For notational simplicity, 
we still denote this temporally smoothed 
prior as $G_t^i$ in subsequent sections.

\begin{figure}[!t]
  \centering
   \includegraphics[width=\linewidth]{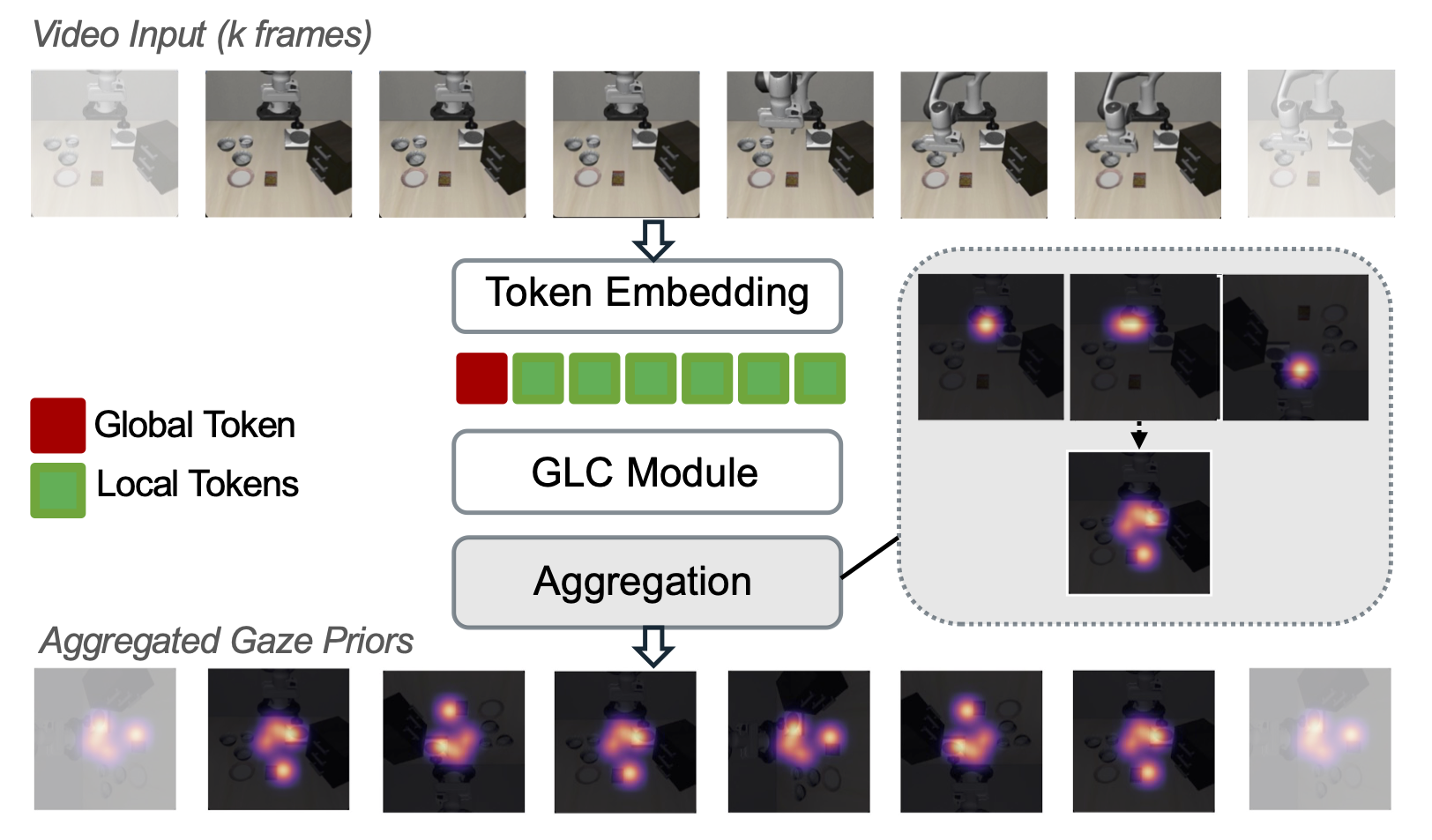}
   \vspace{-2mm}
\caption{\textbf{Temporally Aggregated Gaze Prior Generation.} 
A sequence of $k$ video frames 
is tokenized and processed 
by the GLC \cite{lai2022eye} module, 
which predicts per-frame gaze heatmaps 
using both past and future context. 
These heatmaps are temporally aggregated 
to yield a gaze distribution that captures attention over time and serves as the supervision signal for training- time regularization.}
\vspace{-2mm}
   \label{fig:onecol}
\end{figure}

\subsection{Attention Modulation with Gaze Prior}
\label{sec:attention-modulation}
Having established 
how the gaze prior \(G_t^i\) is derived,
we now describe how it is used 
to modulate the VLA model’s internal attention during training.
Our goal is to align 
the visual regions emphasized by the policy’s vision–language module 
with those that humans naturally attend to 
while completing a task. 
Next, 
we elaborate on a generally applicable framework for attention shaping 
to actionable regions.

\vspace{-2mm}
\paragraph{Model Architecture and Token Interactions}
We introduce the regularization technique 
upon the commonly used Pi-0 architecture \citep{black2024pi0visionlanguageactionflowmodel},
which employs 
a transformer with a causal attention mask 
to govern the information flow 
among different token types.
At each timestep, 
the input sequence to the transformer consists of:
\begin{align*}
X_v^i &\in \mathbb{R}^{N_v \times d} &&\text{: visual tokens from view } i,\\
X_l   &\in \mathbb{R}^{N_l \times d} &&\text{: language tokens},\\
x_p   &\in \mathbb{R}^{1 \times d} &&\text{: proprioceptive token},\\
X_a   &\in \mathbb{R}^{N_a \times d} &&\text{: noisy action tokens (from ground truth)}.
\end{align*}
The causal attention mask 
enforces directional dependencies:
visual-language tokens (\(X_v^i, X_l\)) attend only to VLM tokens,
while action tokens (\(X_a\)) may attend to all preceding tokens.
This creates an information bottleneck 
that preserves the semantics learned by the pretrained vision–language backbone while allowing temporally grounded action prediction.

\vspace{-2mm}
\paragraph{Extracting Model Attention for Regularization}
We extract attention 
from the final transformer layer 
of the vision–language module, 
which provides the most semantically fused 
visual–language features that the action tokens subsequently attend to.
A \emph{global language query} \(Q_{\text{lang}}^{(l)}\) obtained from the language tokens $(X_l)$ summarizes the instruction semantics and attends over the visual tokens.
The resulting attention distribution 
quantifies which image regions 
are most relevant 
to the task represented by the language command:
\begin{equation}
S_t^i = \mathrm{Softmax}\!\left(\frac{Q_{\text{lang}}^{(l)} K_{\text{view}_i}^{(l)\top}}{\sqrt{d}}\right)
    \in \mathbb{R}^{1 \times N_v},
\end{equation}
where \(K_{\text{view}_i}^{(l)}\) are the key vectors corresponding to visual tokens $(X_v^i)$ from view \(i\).
Here $S_t^i[j]$ quantifies the relative importance of the $j^{\text{th}}$ patch in view~$i$ given the language instruction.
This distribution reflects 
how the language query attends to different visual patches, 
effectively capturing the model’s notion 
of task-relevance. 
Next, we illustrate how gaze guides the model's internal attention during training.

\vspace{-2mm}
\paragraph{Gaze-guided Regularization.}
To guide the internal attention 
toward human-like focus patterns, 
we introduce a gaze regularization loss 
based on the Kullback–Leibler divergence between the gaze distribution \(G_t^i\) and the model’s attention distribution \(S_t^i\):
\begin{equation}
\mathcal{L}_{\text{gaze}} = \lambda * D_{\text{KL}}(G_t^i \,\|\, S_t^i).
\label{eq:gaze_loss}
\end{equation}
where $\lambda$ is the coefficient of regularization. This loss acts as a soft alignment term, encouraging but not forcing the model’s attention to mimic human gaze patterns.

The gaze regularization 
operates entirely within the model’s existing causal attention structure.
It modifies the intermediate representations 
of the vision–language module
but does not alter the dependencies enforced by the causal mask.
In effect, 
the regularizer shapes \emph{which} visual–language features the action tokens are encouraged to attend to.
This ensures that gaze supervision improves interpretability and grounding
while maintaining the model’s architecture and deployment efficiency.
We then combine the gaze-guided attention modulation with the standard action-learning objective to form the full training loss.

\subsection{Training Objective and Inference}
\label{sec:training-objective}

The overall training objective 
combines the standard action-learning loss 
with the gaze-regularization term introduced in Eq.~\ref{eq:gaze_loss}.
For each training example \((I_t^{1:n},\, \ell_t,\, q_t,\, A_t^{\ast})\),
the total loss is defined as:
\begin{equation}
\mathcal{L}_{\text{total}}(\theta) 
= \mathcal{L}_{\text{action}}(A_t, A_t^{\ast}) 
+ \lambda\, D_{\mathrm{KL}}\!\left(G_t^i \,\|\, S_t^i\right),
\label{eq:total_loss}
\end{equation}
where \(\mathcal{L}_{\text{action}}\) is the conditional flow matching 
loss used to supervise actions in the VLA 
and the second term acts as a \emph{gaze alignment prior}. 
Specifically,
the first term drives the model 
to reproduce correct robot actions,
while the second introduces a soft inductive bias:
the policy is encouraged 
to prioritize task-relevant visual regions
similar to those attended by humans.
Because both \(S_t^i\) and \(G_t^i\) are normalized distributions over the same set of visual tokens,
their divergence directly measures the alignment between the VLA model's attention and human gaze patterns.

\paragraph{Inference without gaze}
At test time, 
the policy operates entirely without gaze.
The gaze alignment learned during training is implicitly encoded in the model parameters~\(\theta^{\ast}\),
and can be utilized as:
\begin{equation}
A_t = \pi_{\theta^{\ast}}\!\left(I_t^{1:n},\, \ell_t,\, q_t\right).
\label{eq:inference}
\end{equation}
The model thus 
relies solely on its own perception, language, and proprioceptive inputs,
while its internal attention 
naturally reflects human-like focus patterns through the aforementioned training.
This design maintains 
the original real-time efficiency,
enabling robot control 
and embodied interaction tasks
without requiring eye-tracking 
or human gaze estimation at runtime.

\section{Experiments}
\label{sec:experiments}

We evaluate the proposed framework 
by comprehensive experiments 
designed to validate three core hypotheses: 
(1) gaze supervision accelerates learning convergence and improves final task performance across diverse manipulation scenarios; 
(2) the learned attention patterns transfer robustly across task domains; and 
(3) gaze-aligned representations enhance resilience to visual perturbations common in real-world deployment.
Our evaluation spans multiple benchmarks. 
We employ the LIBERO suite \citep{liu2023libero} for comprehensive in-domain analysis, ALOHA-Sim \citep{zhao2023learningfinegrainedbimanualmanipulation} for cross-domain generalization, and OpenVLA \citep{kim2024openvlaopensourcevisionlanguageactionmodel} for architectural transferability. 
Critically, all models operate without gaze input during inference, using only visual, language, and proprioceptive observations to ensure fair comparison.

\subsection{Gaze Regularization Improves Manipulation}

We first investigate whether aligning model attention with human gaze patterns enhances performance on spatially-critical manipulation tasks with the LIBERO-Spatial benchmark, 
which requires precise localization across distinct spatial configurations.

Table~\ref{tab:libero_spatial} demonstrates that gaze regularization yields substantial improvements across most spatial configurations and training stages. 
The regularized model achieves 95.5\% final success compared to 85.9\% for the baseline (9.6 \% gain). 
More revealing is the acceleration of learning: 
at just 10k steps, our method already shows 6.1\% improvement, widening to 7.6\% at 20k steps. 
This early-stage advantage validates our hypothesis that gaze priors provide valuable inductive bias for efficient visual attention allocation, 
enabling models to focus on task-relevant regions rather than discovering them through trial and error.

\begin{table}[!t]
\centering
\caption{
Per-task success rates on LIBERO Spatial ~\citep{liu2023libero} 
The model significantly performs better with spatially-modulated attention.}
\label{tab:libero_spatial}
\resizebox{\columnwidth}{!}{
\begin{tabular}{lcccccc}
\toprule
\multirow{2}{*}{Location of Object} & \multicolumn{3}{c}{\textbf{w Gaze}} & \multicolumn{3}{c}{\textbf{w/o Gaze}} \\
\cmidrule(lr){2-4}\cmidrule(lr){5-7}
 & \textcolor{gray}{10k} & \textcolor{gray}{20k} & \textcolor{gray}{30k} &
   \textcolor{gray}{10k} & \textcolor{gray}{20k} & \textcolor{gray}{30k} \\
\midrule
Between plate and ramekin & 73.3 & 80 & 100 & 70 & 76.7 & 83.3 \\
Next to ramekin           & 60.3 & 71.3 & 100 & 50 & 63.3 & 85.7 \\
Table center              & 76.7 & 80 & 100 & 70 & 80 & 100 \\
On cookie box             & 63.3 & 70.3 & 91.3 & 76.7 & 80 & 100 \\
In cabinet drawer         & 60 & 70 & 73.3 & 43.3 & 50 & 80 \\
On ramekin                & 53.3 & 70 & 100 & 51.3 & 70 & 100 \\
Next to cookie box        & 70 & 90 & 100 & 70 & 90 & 100 \\
On stove                  & 30 & 50 & 90 & 40 & 40 & 90 \\
Next to plate             & 63.3 & 70 & 100 & 20.3 & 36.7 & 50 \\
On wooden cabinet         & 43.3 & 50 & 100 & 40 & 57.7 & 70.3 \\
\midrule
\textbf{Overall Avg.}     & 59.3 & 70.2 & \textbf{95.5} & 53.2 & 62.6 & 85.9 \\
\bottomrule
\end{tabular}
}
\vspace{-2mm}
\end{table}

\vspace{-2mm}
\paragraph{Generalization across Task Domains and Environments}
We next examine whether gaze regularization generalizes beyond spatial manipulation to encompass diverse task types and visual environments. 
We extend our evaluation to the complete LIBERO suite -- including Object manipulation (requiring fine-grained object recognition), Goal-oriented tasks (demanding sequential reasoning), and LIBERO-10 (testing generalization across ten distinct tasks). 
We further validate cross-domain transfer using ALOHA-Sim, presenting fundamentally different visual characteristics and manipulation primitives from LIBERO.

Table~\ref{tab:libero_aloha_results} reveals consistent improvements across all evaluated domains. 
Within LIBERO, gaze regularization yields 8.8\% average improvement at convergence, 
with particularly strong gains on LIBERO-10 (11.8\%), demonstrating enhanced multi-task generalization. 
The temporal progression of improvements -- starting at 4.5\% at 10k steps and expanding to 8.8\% at 30k -- indicates that gaze priors not only accelerate initial learning but continue to provide value throughout training.
The ALOHA-Sim results further validate domain transferability. 
Despite the significant visual and mechanical differences from LIBERO -- including distinct object geometries and manipulation dynamics -- 
our method maintains 4.4\% improvement at convergence. 
The more modest gains reflect ALOHA-Sim's increased task complexity (particularly the challenging peg insertion task), yet the consistent positive transfer demonstrates generalization of the proposed gaze regularization across robotic platforms and environments,
providing a broadly applicable inductive bias rather than dataset-specific heuristics.


\begin{table*}[!ht]
\centering
\small
\caption{Comparison on LIBERO \cite{liu2023libero} and ALOHA Suites \cite{zhao2023learningfinegrainedbimanualmanipulation} (Success Rate \%). 
Each value reports the mean success rate over three seeds at different training steps. 
Columns on the right show the improvement 
of the Gaze-Regularized Model over the Base Model.}
\label{tab:libero_aloha_results}
\begin{tabular}{lcccccccccc}
\toprule
 & \multicolumn{3}{c}{\textbf{w Gaze}} & \multicolumn{3}{c}{\textbf{w/o Gaze}} & \multicolumn{3}{c}{\textbf{$\Delta$ Improvement}} \\
\cmidrule(lr){2-4} \cmidrule(lr){5-7} \cmidrule(lr){8-10}
\ Dataset & \textcolor{gray}{10k} & \textcolor{gray}{20k} & \textcolor{gray}{30k} &
   \textcolor{gray}{10k} & \textcolor{gray}{20k} & \textcolor{gray}{30k} &
   \textcolor{gray}{10k} & \textcolor{gray}{20k} & \textcolor{gray}{30k} \\
\midrule
\textcolor{gray}{\textit{Libero Suite}} \\
\midrule
LIBERO-Spatial & 59.3 & 70.2 & \textbf{95.5} & 53.2 & 62.6 & 85.9 & $\uparrow$ 6.1 & $\uparrow$ 7.6 & $\uparrow$ 9.6 \\
LIBERO-Object  & 76.4 & 86.1 & \textbf{97.3} & 69.5 & 81.2 & 91.7 & $\uparrow$ 6.9 & $\uparrow$ 4.9 & $\uparrow$ 5.6 \\
LIBERO-Goal    & 72.8 & 83.5 & \textbf{92.6} & 66.9 & 77.4 & 84.3 & $\uparrow$ 5.9 & $\uparrow$ 6.1 & $\uparrow$ 8.3 \\
LIBERO-10      & 41.7 & 58.3 & \textbf{77.9} & 42.5 & 53.8 & 66.1 & $\downarrow$ 0.8 & $\uparrow$ 4.5 & $\uparrow$ 11.8 \\
\textbf{Average} & 62.6 & 74.5 & \textbf{90.8} & 58.1 & 68.8 & 82.0 & $\uparrow$ 4.5 & $\uparrow$ 5.7 & $\uparrow$ 8.8 \\
\midrule
\textcolor{gray}{\textit{Aloha-Simulation Gym-Aloha}}\\
\midrule
Transfer Cube  & 40.0 & 65.0 & \textbf{77.5} & 36.2 & 58.8 & 72.5 & $\uparrow$ 3.8 & $\uparrow$ 6.2 & $\uparrow$ 5.0 \\
Peg Insertion   & 0.0 & 12.5 & \textbf{18.8} & 0.0 & 8.8 & 15.0 & 0 & $\uparrow$ 3.7 & $\uparrow$ 3.8 \\
\textbf{Average} & 20 & 38.8 & \textbf{48.2} & 18.1 & 33.8 & 43.8 & $\uparrow$ 1.9 & $\uparrow$ 5 & $\uparrow$ 4.4 \\
\bottomrule
\end{tabular}
\end{table*}

\vspace{-2mm}
\paragraph{Architectural Transferability with OpenVLA}
A critical test of our framework's generality lies in its transferability across different model architectures. 
While our primary experiments utilize Pi-0, practical deployment requires methods that enhance existing systems without architecture-specific modifications. 
We therefore evaluate whether gaze regularization maintains its effectiveness when applied to OpenVLA~\cite{kim2024openvlaopensourcevisionlanguageactionmodel}, a structurally distinct VLA model,
thus, validating if our method 
operates at a fundamental level than exploiting architecture-specific properties. 


\begin{table}[!t]
\centering
\small
\begin{scriptsize}
    \caption{
    Comparison of Base 
    and Gaze-Regularized models 
    with OpenVLA~\cite{kim2024openvlaopensourcevisionlanguageactionmodel}. 
    Our proposed method 
    achieves higher performance 
    even under a different architectural setup.}
    \vspace{-2mm}
\label{tab:open_vla}
\begin{tabular}{lccc}
\toprule
\ Dataset & \textbf{w/o Gaze} & \textbf{w Gaze} & \textbf{$\Delta$ Improvement} \\
\midrule
LIBERO-Spatial & 76.0 & \textbf{82.2} & $\uparrow$ 6.2 \\
LIBERO-Object  & 79.5 & \textbf{86.1} & $\uparrow$ 6.6 \\
LIBERO-Goal    & 72.5 & \textbf{76.8} & $\uparrow$ 4.3 \\
LIBERO-10      & 45.9 & \textbf{51.5} & $\uparrow$ 5.6 \\
\midrule
\textbf{Overall Avg.} & 68.5 & \textbf{74.2} & $\uparrow$ 5.7 \\
\bottomrule

\end{tabular}
\end{scriptsize}
\vspace{-2mm}
\end{table}

Table~\ref{tab:open_vla} presents the comparative results when both baseline and gaze-regularized OpenVLA variants are trained identically on the LIBERO suite. 
The gaze-regularized model achieves consistent improvements of 4 - 6\% across all task categories, 
with an overall gain of 5.7\%. 
The consistent improvements 
across all LIBERO suites demonstrate 
that even models with strong multimodal pretraining benefit from explicit gaze supervision. 
These architectural transfer results, combined with our cross-domain validation, establish gaze regularization as a model-agnostic enhancement 
rather than architecture-specific mechanisms, and since our approach is meant to be modular, it can easily be integrated into existing architectures.
 \vspace{-2mm}
\paragraph{Implementation on Real-Life Robot}
\label{sec:real-life-robot}
We further validated our approach by deploying it on a physical robotic system across three manipulation tasks with varying complexity. These tasks were designed to test two different time horizons: short-horizon tasks requiring a single action sequence, and a longer-horizon task requiring sequential actions. For the short-horizon category, we included two tasks: (1) picking up a cube and placing it on a plate, and (2) picking up a cup and placing it in a container. For the longer-horizon task, we challenged the robot to pick up multiple cups one by one and place each in the container, testing its ability to maintain attention and execute repeated actions. We evaluated both a baseline policy and our gaze-regularized policy on these tasks. The results demonstrate consistent improvements with our approach: an 8\% increase in success rate across the short-horizon tasks, and a 10\% improvement in success rate for the longer-horizon task. These findings confirm that the benefits of gaze regularization transfer from simulation to real-world robotic manipulation, with even greater gains observed in more complex, multi-step scenarios. 


\begin{table}
\centering
\small
\begin{scriptsize}
\caption{
Comparison of Base and Gaze-Regularized models on real-world tasks using Pi-0.}
\vspace{-2mm}
\label{tab:real-world}
\begin{tabular}{lccc}
\toprule
\textbf{Task} & \textbf{Steps} & \textbf{Base Model} & \textbf{Gaze Model} \\
\midrule
Place the cube & 20,000 & 4\% & 6\% \\
on the plate & 40,000 & 32\% & 44\% \\
\midrule
Pick cup and place & 20,000 & 24\% & 28\% \\
it in container & 40,000 & 64\% & 72\% \\
\midrule
Pick multiple cups & 20,000 & 5\% & 5\% \\
and place in container & 40,000 & 30\% & 40\% \\
\bottomrule
\end{tabular}
\end{scriptsize}
\vspace{-2mm}
\end{table}

\subsection{Ablation Studies}
\label{sec:ablation_studies}
We now dissect the framework's key design choices through systematic ablation studies. 
These experiments isolate  critical components that govern the method's success:the strength of regularization during training and the robustness of learned representations under visual degradation, 
providing 
practical guidance for implementation.

\vspace{-2mm}
\paragraph{Sensitivity to Gaze Regularization Scale}
\label{sec:gaze_lambda}
The regularization coefficient $\lambda$ in Equation~\ref{eq:total_loss} controls the balance between action learning and gaze alignment, 
determining whether human attention patterns serve as gentle guidance or strict constraints. 
This parameter shapes how the model integrates perceptual priors with task-specific learning. 
We investigate three distinct regularization regimes to identify the optimal balance: weak regularization (0.001) that provides soft bias, moderate regularization (0.01) that more strongly influences attention, and strong regularization (10) that heavily prioritizes gaze alignment.

Table~\ref{tab:gaze_lambda} demonstrates a clear optimal range, with weak regularization (0.001) achieving the highest performance at 90.8\% 
average success. 
Moderate regularization maintains baseline-comparable performance at 82.2\%, 
while strong regularization catastrophically degrades to 41.6\%. 
The success of weak regularization confirms that human gaze functions most effectively as a soft inductive bias rather than a hard constraint. 
By maintaining low regularization strength, the model benefits from the statistical tendencies of human attention while preserving flexibility to discover task-optimal patterns that may occasionally deviate from human gaze. 
This calibration study establishes that gaze regularization succeeds precisely because it guides without constraining, accelerating the discovery of task-relevant features while allowing the model to refine these patterns based on action outcomes. 

\begin{table}[h]
\centering
\small
\caption{Effect of gaze-regularization strength ($\lambda$) on task success rate (\%). 
The model shows improved performance when gaze is used as a soft prior rather than a hard constraint.}
\vspace{-2mm}
\begin{scriptsize}
\begin{tabular}{lcccc}
\toprule
& &\multicolumn{3}{c}{\textbf{Regularization Scale}} \\
\cmidrule(lr){3-5}
Suite & Baseline & \textcolor{gray}{Low}  & \textcolor{gray}{Moderate}  & \textcolor{gray}{High}  \\
\midrule
LIBERO-Spatial & 85.9 & \textbf{95.5} & 88.4 & 44.2 \\
LIBERO-Object  & 91.7 &\textbf{97.3} & 90.8 & 50.6 \\
LIBERO-Goal    & 84.3 &\textbf{92.6} & 85.1 & 41.7 \\
LIBERO-10      & 66.1 &\textbf{77.9} & 64.4 & 30.1 \\
\midrule
\textbf{Overall Avg.} & 82 & \textbf{90.8} & 82.2 & 41.6 \\
\bottomrule
\end{tabular}
\end{scriptsize}
\label{tab:gaze_lambda}
\vspace{-2mm}
\end{table}

\paragraph{Alignment of predicted gaze with ground truth gaze }
\label{sec:aligntment_gaze}
To evaluate how well our synthetic gaze predictions match real human gaze patterns, we conducted a validation study using an eye tracking device we borrowed. We recruited participants and had them watch simulation videos while following specific task instructions, capturing their actual eye movements as ground truth data. We then compared our model's predicted gaze heatmaps against this real eye tracking data using region-level Intersection over Union (IoU). Specifically, we identified the top-k regions where humans looked most frequently based on eye tracking and calculated their overlap with our model's top-k predicted gaze regions. The results demonstrate strong alignment between synthetic and real gaze, with synthetic heatmaps achieving 68.6\% mean IoU for the top-32 regions and 82.3\% mean IoU for the top-64 regions indicating our synthetic gaze predictions closely mirror where humans actually look when watching simulation videos.

\begin{table}[!t]
\centering
\caption{
Performance comparison under different visual perturbations when noise or visual degradations are simulated.
Results are reported across three LIBERO benchmarks \cite{liu2023libero}. 
}
\label{tab:robustness_results}
\resizebox{\columnwidth}{!}{
\begin{tabular}{lcccccc}
\toprule
\multirow{2}{*}{\textbf{Perturbation}} &
\multicolumn{2}{c}{\textbf{LIBERO-Spatial}} &
\multicolumn{2}{c}{\textbf{LIBERO-Object}} &
\multicolumn{2}{c}{\textbf{LIBERO-Goal}} \\
\cmidrule(lr){2-3}\cmidrule(lr){4-5}\cmidrule(lr){6-7}
 &\textcolor{gray}{w/o Gaze} & \textcolor{gray}{w Gaze} & \textcolor{gray}{w/o Gaze} & \textcolor{gray}{w Gaze} & \textcolor{gray}{w/o Gaze} & \textcolor{gray}{w Gaze} \\
\midrule
No Perturbation   & 85.9 & \textbf{95.5} & 91.7 & \textbf{97.3} & 84.3 & \textbf{92.6} \\
Lighting Variation & 77.2 & \textbf{89.1} & 84.5 & \textbf{92.8} & 80.4 & \textbf{89.7} \\
Camera Noise       & 82.1 & \textbf{91.3} & 85.8 & \textbf{93.5} & 79.6 & \textbf{88.9} \\
\bottomrule
\end{tabular}
}
\vspace{-3mm}
\end{table}

\vspace{-2mm}
\paragraph{Robustness under Visual Perturbations}
\label{sec:robustness_perturb}

Real-world robots must operate under visual conditions that deviate from training environments—variable lighting, sensor noise, and optical distortions are the norm, not exceptions. A critical question is whether gaze-regularized models, with their focused attention patterns, maintain advantages when visual inputs are corrupted, or whether this focus becomes a liability when those regions are degraded. Table~\ref{tab:robustness_results} reveals that gaze regularization amplifies its advantages under such perturbations. Under lighting variations, the performance gap widens across all benchmarks—LIBERO-Spatial shows an 11.9\% advantage (89.1\% vs 77.2\%) compared to 9.6\% under normal conditions
Similarly, under camera noise, gaze-regularized models maintain strong advantages across Spatial (91.3\% vs 82.1\%), Object (93.5\% vs 85.8\%), and Goal (88.9\% vs 79.6\%) tasks, demonstrating resilience to pixel-level corruption by attending to semantic features that persist despite sensor noise. Combined with its training-only implementation and architectural flexibility, this robustness positions gaze regularization as a practical enhancement for VLA-based systems operating in unstructured environments

\section{Conclusion}
\label{sec:conclusion}

We present a gaze-regularized training-only framework 
that addresses the attention allocation challenge 
in VLA models. 
our approach achieves consistent performance improvements across diverse benchmarks without requiring architectural modifications or inference dependencies. 
While our current implementation 
leverages synthetic gaze from pretrained models, 
future integration of real eye-tracking data from expert demonstrations could further strengthen these benefits. 
The framework's training-only implementation 
enables immediate deployment as a practical enhancement for existing robotic systems, 
with our finding that a soft regularization performs optimally revealing human attention functions best as flexible guidance. 
As autonomous systems increasingly operate in human environments, 
incorporating human perceptual strategies through gaze supervision offers a principled approach to achieving more capable and interpretable robotic manipulation. 
Our results establish that bridging human cognitive patterns with machine learning represents an essential pathway toward human-level performance in complex real-world tasks.

{
    \small
    \bibliographystyle{ieeenat_fullname}
    \bibliography{main}
}


\maketitlesupplementary
\setcounter{page}{1}
\appendix

This supplementary document provides extended methodological details, additional ablations and implementation clarifications to support the claims made in the main paper. The structure is as follows:

\begin{itemize}
    \item Appendix A: Notation Table
    \item Appendix B: Expanded Methodological Clarifications
    \item Appendix C: Additional Attention - Gaze Alignment Evidence
    \item Appendix D: Synthetic Gaze Reliability and Ablations
    \item Appendix E: Other Experiments
    \item Appendix F: Pseudo-code and Reproducibility Details
    \item Appendix G: Summary of Additions and Discussion
\end{itemize}

\begin{table*}[t]
\centering
\caption{Summary of key notations used in gaze-to-attention regularization and VLA token interactions.}
\vspace{1mm}
\begin{tabular}{ll}
\toprule
\textbf{Symbol} & \textbf{Description} \\
\midrule
$t$ & Timestep index of the current observation \\
$i$ & Camera/view index \\
$N_v$ & Number of visual tokens (e.g., $16 \times 16 = 256$) \\
$P$ & Patch grid dimension (e.g., $P=16$) \\[1mm]

$X_l \in \mathbb{R}^{N_l \times d}$ & Language token sequence \\
$X_v^i \in \mathbb{R}^{N_v \times d}$ & Visual tokens from camera view $i$ \\
$Q_{\text{lang}}^{(l)} \in \mathbb{R}^{1 \times d}$ & Global query summarizing language semantics \\
$K_{\text{view}_i}^{(l)} \in \mathbb{R}^{N_v \times d}$ & Key vectors for visual tokens from view $i$ \\[1mm]

$H_{t}^i \in \mathbb{R}^{H_g \times W_g}$ & Predicted gaze heatmap for view $i$ at time $t$ \\
$\tilde{H}_{t}^i$ & Temporally aggregated gaze heatmap centered at $t$ \\
$G_{t}^i \in \mathbb{R}^{N_v}$ & Patch-level gaze distribution for view $i$ \\[1mm]

$S_{t}^i \in \mathbb{R}^{N_v}$ & Model’s spatial attention over visual tokens \\
$D_{\mathrm{KL}}(G_{i,t} \,\|\, S_{i,t})$ & KL divergence measuring gaze–attention alignment \\[1mm]

$I_{t}^i$ & RGB frame from view $i$ at time $t$ \\
$\ell_t$ & Tokenized language instruction \\
$q_t$ & Proprioceptive observation at time $t$ \\[1mm]

$A_t$ & Predicted short-horizon action sequence \\
$A_t^\ast$ & Ground-truth action sequence \\[1mm]

$\lambda$ & Gaze-regularization weighting coefficient \\
$T$ & Temporal aggregation window size for gaze \\
\bottomrule
\end{tabular}
\label{tab:notation_summary}
\end{table*}


\section{Notation and Symbol Table}

To improve clarity and provide a quick reference for readers, we summarize the key notations used throughout the paper and supplementary material. These symbols cover visual tokens, patch grids, gaze heatmaps, attention matrices, and their corresponding distributions.

\section{Expanded Methodological Clarifications}

In this section, we provide additional details on how gaze supervision is integrated into the VLA architecture. We first clarify how spatial attention is extracted and regularized, then discuss the properties and reliability of the predicted gaze signals used throughout our experiments. These clarifications are intended to make the connection between model internals, gaze priors, and action prediction more explicit than in the main paper.

\paragraph{Constructing a Singular Global Query from Language Tokens.}
To obtain a unified representation of the instruction, we collapse the sequence of language embeddings into a single global query vector. This can be implemented through simple pooling, a learned linear projection, or a lightweight attention-based aggregator; in our implementation, a simple projection maps the full language-token sequence $\{X_{l}^{(1)}, \dots, X_{l}^{(N_l)}\}$ into a compact semantic vector $Q_{\text{lang}}$. This vector captures the dominant intent of the instruction and serves as a query over the visual scene. 

\paragraph{Detailed Attention Extraction}

Our approach introduces gaze-guided supervision into the VLA model by regularizing its \emph{internal spatial attention} during training. Since robots do not possess an innate mechanism analogous to human eye-gaze, the goal is to endow the policy with a learned surrogate of gaze i.e a structured  prior that encourages the transformer to focus on task-relevant regions during manipulation.

The spatial attention regularized in our framework emerges from the interaction between the vision and language streams in the final transformer layer of the VLA backbone. The language encoder first produces a sequence of instruction tokens 
\( X_l \in \mathbb{R}^{N_l \times d} \),
which are aggregated through a learned projection to form a global query vector (as mentioned in the previous paragraph)
\( Q^{(l)}_{\text{lang}} \).
This query functions as a compact representation of the semantics of the task instruction.

For each camera view \(i\), the visual encoder outputs a set of tokens
\( X_v^i \in \mathbb{R}^{N_v \times d} \),
which are linearly projected to key vectors
\( K_{\text{view}_i}^{(l)} \),
following the standard attention formulation established in~\citep{vaswani2023attentionneed}. 
The resulting cross-attention captures the degree to which each visual patch is relevant to the language instruction:

\[
S_t^i 
= \operatorname{Softmax}\!\left(
\frac{
Q^{(l)}_{\text{lang}} {K_{\text{view}_i}^{(l)}}^{\top}
}{\sqrt{d}}
\right)
\in \mathbb{R}^{1 \times N_v}.
\]

This attention distribution quantifies the importance assigned to each visual token when interpreting the task instruction. We extract the attention distribution specifically from the \textbf{final vision--language transformer layer}, for two reasons:

\begin{enumerate}
    \item \textbf{Semantic maturity.}  
    Late transformer layers might contain the most semantically integrated features, combining spatial, linguistic, and contextual cues.
    
    \item \textbf{Action relevance.}  
    In Pi-0 and other VLA architectures, the action tokens attend to the fused representations produced by the final vision--language layer. Thus, regularizing this layer directly shapes the perceptual information used for motor prediction.
\end{enumerate}

This design parallels observations from prior work such as~\citep{panigaze}, which shows that late-layer attention better reflects task-relevant perceptual cues. However, unlike prior methods, our approach applies this principle to \textbf{robotic control settings}, where attention not only guides prediction but directly influences action generation.

Aligning this spatial attention with human gaze priors yields an inductive bias that is both \emph{compact} and \emph{action-grounded}. This approach mirrors core aspects of human behavior: just as humans internalize a rich understanding of a scene--fusing visual cues with linguistic and contextual knowledge before executing a precise motor action, our method regularizes the model's final representations to guide its decisions. Consequently, the policy is encouraged to mirror the fixation and information-gathering strategies humans employ before and during manipulation.

\subsection{Reliability of Predicted Gaze}

Because robotic datasets rarely include human eye-tracking labels, we employ \emph{synthetic gaze} generated by pretrained gaze-estimation networks. Among existing models, we adopt the Global--Local Correlation (GLC) network~\citep{lai2022eye} due to a combination of temporal fidelity, robustness, and strong performance on egocentric video tasks.

\vspace{-3mm}
\paragraph{Temporal Sensitivity.}
Human gaze during manipulation is inherently dynamic: fixations shift in anticipation of upcoming hand movements. GLC explicitly models these temporal dependencies by processing short clips rather than single frames, producing gaze heatmaps informed by both past and future context. This confers a key advantage over earlier single-frame models such as DeepGaze~\citep{kümmerer2015deepgazeiboosting}, although DeepGaze and it's new variants show great performance in tasks which require a scanning pattern over a static scene, and in the future, this ability can be leveraged to make our method even better.

\paragraph{Strong Performance in Manipulation-like Settings.}
GLC achieves high accuracy on egocentric and hand--object interaction datasets, which share structural similarities with robotic manipulation scenes (clutter, hand presence, fine-grained object interactions). These properties make GLC particularly suitable for generating gaze priors for multi-view robotic datasets. In the future, curated teleoperated datasets with ground-truth gaze could further improve interpretability and accuracy by providing real human fixation patterns rather than synthetic estimates.

\vspace{-2mm}
\paragraph{Ablations on Gaze Quality.}
To verify that performance improvements stem from meaningful gaze characteristics rather than incidental regularization, we perform additional robustness experiments (see later appendices):
\begin{itemize}
    \item \textbf{DeepGaze comparison:} replacing GLC with DeepGaze reduces performance, indicating that accurate spatial structure of gaze is important.
    \item \textbf{Uniform Gaze:} by equally dividing attention across all the patches, the benefits are not seen anymore, confirming that only \emph{structured} gaze provides useful supervision.

\end{itemize}

While synthetic gaze is inherently an approximation of true human fixation behavior, our experiments demonstrate that it provides a powerful supervisory signal for shaping transformer attention. We view our results as an initial bound on the benefits achievable with real eye-tracking, and anticipate even greater gains as future teleoperation datasets incorporate true human gaze measurements.

\begin{table}[!t]
\centering
\caption{
Per-task success rates on LIBERO Spatial~\citep{liu2023libero} at 30k training steps.
We compare the baseline model, our gaze-regularized model, a DeepGaze-based gaze variant, and a uniform-distribution variant.}
\label{tab:libero_spatial_30k_full}
\resizebox{\columnwidth}{!}{
\begin{tabular}{lcccc}
\toprule
\multirow{2}{*}{Location of Object} 
& \textbf{w Gaze} 
& \textbf{DeepGaze} 
& \textbf{w/o Gaze} 
& \textbf{Uniform} \\
\cmidrule(lr){2-5}
 & \textcolor{gray}{30k} & \textcolor{gray}{30k} & \textcolor{gray}{30k} & \textcolor{gray}{30k} \\
\midrule
Between plate and ramekin    & 100   & 85.7 & 83.3 & 69.7 \\
Next to ramekin              & 100   & 86.7 & 85.7 & 59.7 \\
Table center                 & 100   & 100  & 100  & 80.3 \\
On cookie box                & 91.3  & 100  & 100  & 79.3 \\
In cabinet drawer            & 73.3  & 82.0 & 80   & 39.3 \\
On ramekin                   & 100   & 100  & 100  & 50.7 \\
Next to cookie box           & 100   & 100  & 100  & 50.3 \\
On stove                     & 90    & 91.0 & 90   & 10.3 \\
Next to plate                & 100   & 55.0 & 50   & 70.7 \\
On wooden cabinet            & 100   & 73.3 & 70.3 & 60.3 \\
\midrule
\textbf{Overall Avg.}        
& \textbf{95.5} 
& {86.3} 
& 85.9 
& 57.1 \\
\bottomrule
\end{tabular}
}
\vspace{-2mm}
\end{table}

\begin{figure}[!t]
  \centering
   \includegraphics[width=\linewidth]{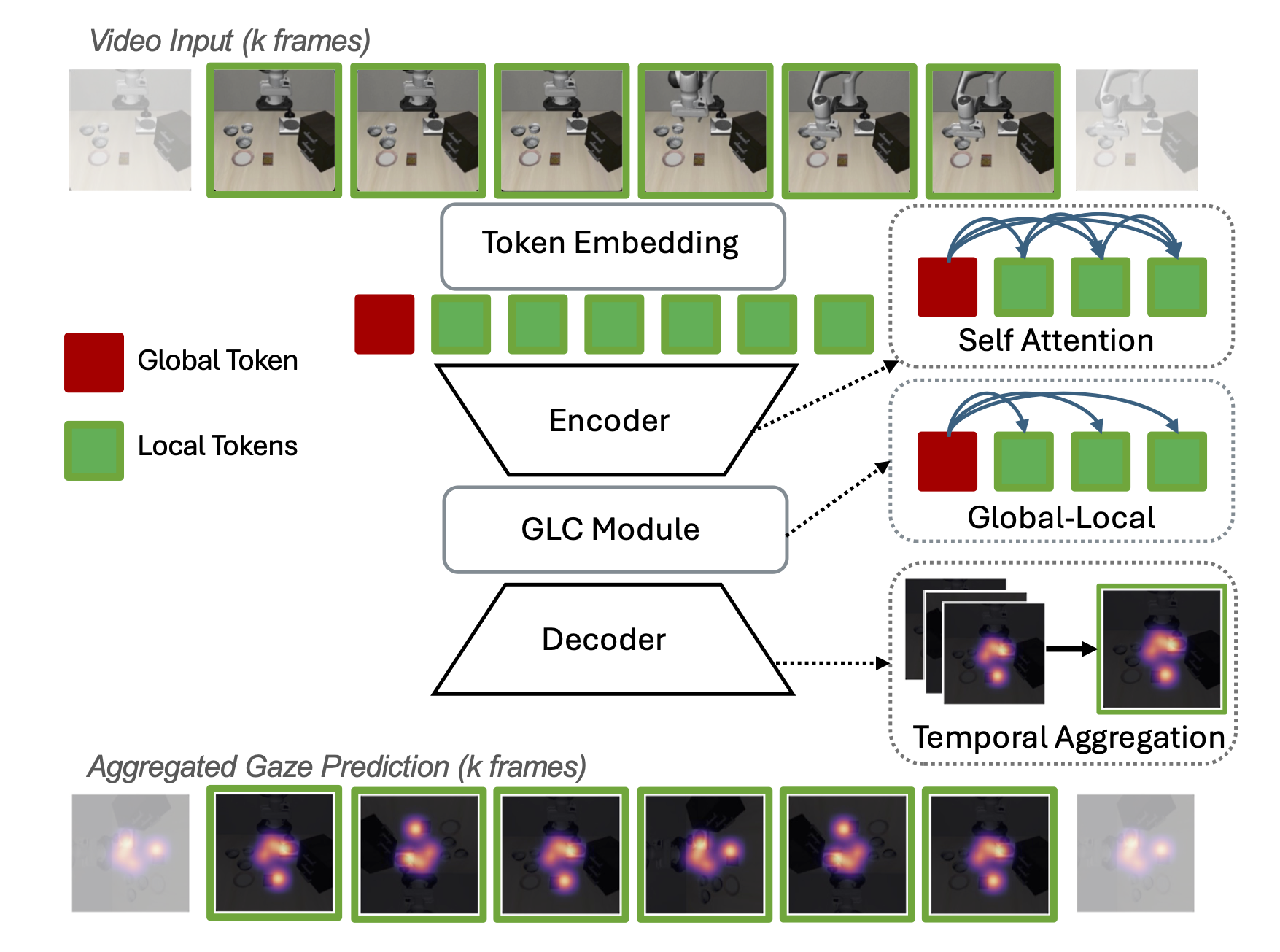}
   \vspace{-2mm}
\caption{\textbf{Closer look at Gaze Prior Generation} 
A sequence of $k$ video frames 
is tokenized and processed 
by the GLC \cite{lai2022eye} module, 
where it utilizes global tokens (derived from the sequence) and local tokens, and undergoes self attention as well as Global-Local Correlation to then predict per-frame gaze heatmaps. 
These heatmaps are temporally aggregated 
to yield a gaze distribution that captures attention over time and serves as the supervision signal for training- time regularization.}
\vspace{-4mm}
   \label{fig:aggregation_extended}
\end{figure}

\section{Attention--Gaze Alignment Evidence}

Beyond task success rates, a core claim of our work is that gaze regularization shapes the model’s internal attention to better reflect human fixation patterns. In this section, we first introduce a quantitative Top-$k$ overlap metric to measure alignment between model attention and gaze distributions, and then provide additional qualitative visualizations to illustrate how this alignment manifests across tasks, viewpoints, and time.

\subsection{Top-$k$ Attention--Gaze Overlap Metrics}

A central question in evaluating our framework is whether gaze regularization meaningfully shifts the model's internal attention toward human fixation patterns. While qualitative visualizations already suggest improved alignment, we seek a more rigorous quantitative measure. To this end, we compute a \emph{Top-$k$ attention--gaze overlap} metric that assesses how frequently the model's most attended patches coincide with regions prioritized by human gaze. For our experiment, we use a value of k=10.

\paragraph{Metric Definition.}
For each view $i$ at time $t$, let
$S_t^i \in \mathbb{R}^{N_v}$ denote the model's spatial attention distribution and
$G_t^i \in \mathbb{R}^{N_v}$ denote the gaze-derived patch-level distribution.
We identify the indices of the model's $k$ highest-attended patches:
\[
\mathcal{T}_k(S_t^i) = \text{Top-}k(S_t^i).
\]
We then compute the total gaze mass contained within these patches:
\[
\text{Overlap}_k(t,i)
= 
\sum_{j \in \mathcal{T}_k(S_t^i)} G_{t,j}^i.
\]

This yields a score in $[0,1]$, where a value of $1$ indicates that all gaze probability lies within the model's top-$k$ attended patches, and $0$ indicates complete misalignment.

\begin{figure*}[!t]
  \centering
   \includegraphics[width=\linewidth]{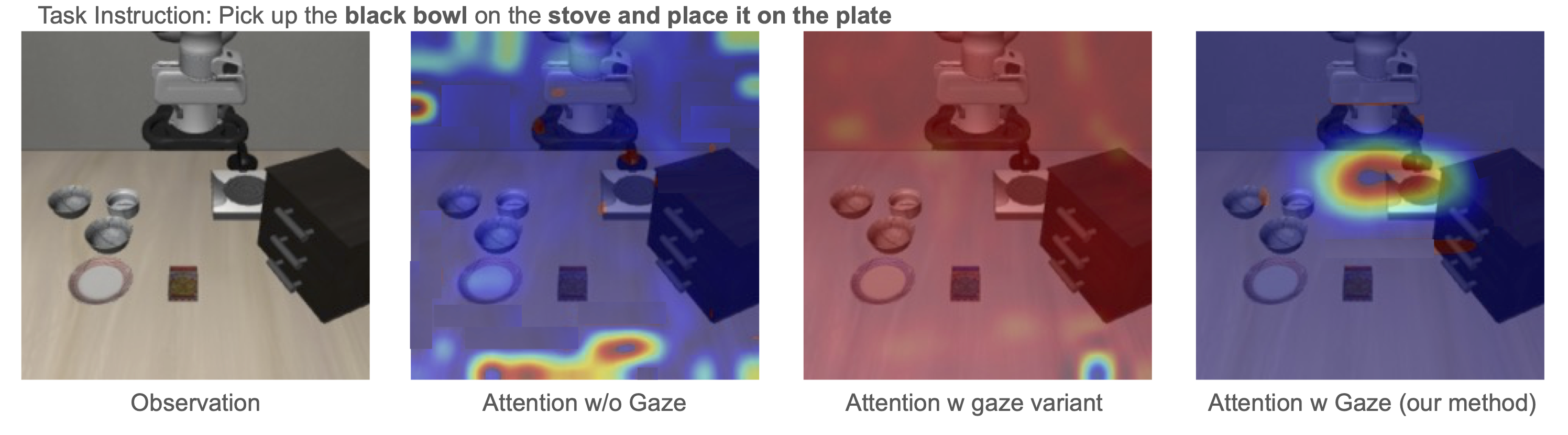}
   \vspace{-2mm}
\caption{\textbf{Additional Visualisations of Attention.} 
Given the input observation, we show the spatial attention from the baseline model (second), the attention obtained when a perturbed gaze variant is used (third, corresponding to Table~\ref{tab:libero_spatial_30k_full}), and finally the sharper, task-relevant attention produced by our gaze-regularized model (fourth).}
\vspace{-2mm}
   \label{fig:attn_vis}
\end{figure*}

We observe a substantial improvement in overlap after applying gaze regularization. For example, at $k{=}10$, the baseline model achieves an average overlap of $19\%$, whereas the gaze-regularized model achieves $51\%$. The relative improvement  indicates that the regularized model attends more sharply to the most gaze-salient regions, as shown in Figure~\ref{fig:attn_vis}.

 \subsection{Attention Map Visualizations}
To complement the Top-$k$ quantitative analysis, we include an additional qualitative comparison of spatial attention maps across three settings: the baseline model (no gaze), a model trained with a gaze variant, and our proposed gaze-regularized model. This visualization clearly highlights the characteristic differences produced by each training scheme.

Across all views shown in Figure~\ref{fig:attn_vis}, we observe that the baseline model exhibits diffuse and spatially inconsistent attention, often spreading mass across irrelevant background regions. Using an uniform gaze prior produces diffused attention as well , and still lacks strong task grounding. In contrast, our method produces sharply localized and semantically aligned attention, focusing on regions directly relevant to the instructed manipulation.

These visual patterns are consistent with and supportive of the Top-$k$ overlap results reported earlier: the gaze-regularized model’s attention aligns more closely with human fixation structure, reflecting a more task-aware perceptual representation.

\subsection{Attention Modulation using average representation of all layers}

In the main paper, we regularize the spatial attention extracted from the \emph{final} vision--language transformer layer. This design choice is motivated by the fact that the last layer contains the most semantically integrated features, and its attention maps directly govern the information available to the action tokens. A natural question, however, is whether distributing gaze supervision across \emph{all} layers might further improve performance or stability.

To investigate this, we consider a variant in which we first compute the attention distribution at each transformer layer, then average these distributions across depth, and finally apply the gaze regularization loss to this layer-averaged attention. Intuitively, this variant encourages gaze-aligned information flow throughout the entire network, rather than only at the last layer.

Table~\ref{tab:libero_all_layers} reports per-task success rates on LIBERO-Spatial when regularizing this averaged attention across all layers. We observe that this variant achieves competitive performance when compared to the baseline model across most spatial configurations and training checkpoints. At the same time, the results support our design choice in the main paper: concentrating gaze supervision on the final vision--language layer provides a larger increase in accuracy while incurring no additional overhead from multi-layer aggregation.

\begin{table}[!t]
\centering
\caption{
Per-task success rates on LIBERO Spatial with regularization applied to all layers.
The model shows competitive performance with comprehensive regularization.}
\label{tab:libero_all_layers}
\begin{tabular}{lcccc}
\toprule
\multirow{2}{*}{Location of Object} & \multicolumn{3}{c}{\textbf{w Gaze (All Layers)}} \\
\cmidrule(lr){2-4}
& \textcolor{gray}{10k} & \textcolor{gray}{20k} & \textcolor{gray}{30k} \\
\midrule
Between plate and ramekin & 65.0 & 75.0 & 90.3 \\
Next to ramekin & 55.0 & 70.0 & 89.7 \\
Table center & 70.0 & 85.0 & 100.0 \\
On cookie box & 58.3 & 65.0 & 70.3 \\
In cabinet drawer & 43.3 & 56.3 & 60.7 \\
On ramekin & 48.3 & 65.0 & 99.7 \\
Next to cookie box & 65.0 & 85.0 & 100.0 \\
On stove & 25.0 & 45.0 & 79.3 \\
Next to plate & 56.7 & 70.0 & 99.3 \\
On wooden cabinet & 38.3 & 55.0 & 80.3 \\
\midrule
\textbf{Overall Avg.} & 52.5 &  67.1& \textbf{87} \\
\bottomrule
\end{tabular}
\vspace{-2mm}
\end{table}

\begin{figure*}[!t]
  \centering
   \includegraphics[width=\linewidth]{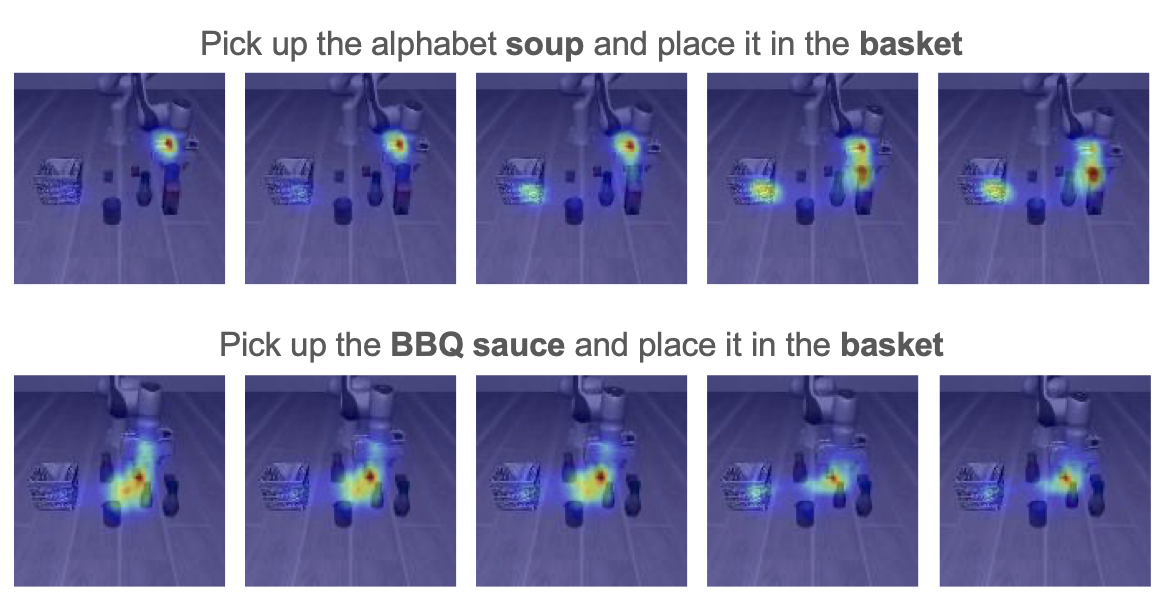}
   \vspace{-2mm}
\caption{\textbf{ Reliability of Synthetic Gaze on Simulation Videos} 
Given the input task, we show the the predicted gaze is accurate and even on similar visual settings, produces different gaze results depending on the language instruction. The model utilizes a temporal sequence of frames, rather than a single frame, and then computes the gaze prediction thus the prediction occurs due to the conditioning through the global context by operating on a sequence of frames}
\vspace{-2mm}
   \label{fig:gaze_reliability}
\end{figure*}

\subsection{Task-Conditioned Gaze and Language-Conditioned VLM Attention}
\label{sec:task-condition-gaze}
We establish that informative gaze during manipulation is task-dependent and that different language instructions can induce different gaze patterns. In our framework, gaze is predicted from short temporal sequences rather than single images, allowing the gaze model to exploit action progression and implicit task context. Since the model was trained using data from a task-driven setting rather than free viewing, the predicted aggregated gaze yields top-down, task-driven attention rather than bottom-up saliency. From Figure ~\ref{fig:gaze_reliability}, we can also see that the temporal processing of a sequence of frames provides the task context, and hence produces different gaze results for different task instructions, even under similar settings.

Human visual attention in this work refers specifically to egocentric, action-oriented, top-down gaze during object manipulation. Fixations anticipate contact regions, targets, and task-relevant spatial relations rather than free-viewing or social gaze. Temporal aggregation further captures anticipatory fixations that precede motor execution, consistent with findings in the action-perception literature. 

Crucially, language is explicitly incorporated through the VLM attention we regularize. The attention map is extracted using a global language token (derived from the instruction) as the query over visual tokens, making it inherently language and task-conditioned. Gaze therefore does not replace task reasoning; it provides a soft spatial prior that biases where a task-aware VLM attends. Regularization is applied as a soft constraint, and gaze–attention overlap is partial (i.e., 51\% top-10 overlap in our method vs. 19\% in baseline), not enforced to be identical.

If temporally predicted, image-only gaze were incompatible with language conditioned attention, performance would degrade on tasks with similar observations but different instructions (e.g., LIBERO). Instead, we observe consistent improvements across such settings, indicating that temporally predicted gaze complements, rather than conflicts with, task-aware VLM attention.

\section{Other Experiments}

Beyond the standard evaluation settings presented in the main paper, it is important to understand whether gaze regularization provides benefits under conditions that more closely resemble real-world deployment. Robots operating outside controlled laboratory environments routinely face perturbations in both visual observations and task instructions. In this appendix, we therefore expand our analysis to two additional scenarios: (i) linguistic perturbations that modify the phrasing of task instructions, and (ii) cross-viewpoint degradation where one of the camera inputs becomes unavailable. Together, these experiments shed light on the robustness and generalization properties of gaze-regularized VLA models.

\subsection{Perturbations in Language Prompts as Task Distractors}

While Section~\ref{sec:ablation_studies} introduces visual perturbations, linguistic perturbations can also serve as practical task distractors. Natural language in the real world is rarely fixed: users may rephrase commands, substitute synonyms, or give instructions with subtle differences in wording. To simulate such conditions, we manually replaced verbs in the LIBERO-Spatial instruction set with alternatives such as \emph{grab}, \emph{retrieve}, or \emph{lift} in place of the canonical \emph{pick}. All prompts were kept similar in length to avoid introducing length-based biases.

We then compare model performance under these instruction variations for both the baseline (without gaze regularization) and our gaze-regularized approach in Table~\ref{tab:prompt_distractors}. The drop in performance is similar across both models, but the gaze-regularized approach still performs better overall, even when the linguistic phrasing deviates from the  distribution seen during training.

\begin{table}[!t]
\centering
\caption{
Per-task success rates on LIBERO Spatial under prompt distractors (e.g., replacing ``pick'' with ``grab'', ``lift'', etc.). 
Both the baseline and gaze-regularized models exhibit performance degradation, but the gaze model remains more robust.}
\label{tab:prompt_distractors}
\resizebox{\columnwidth}{!}{
\begin{tabular}{lcc}
\toprule
\multirow{2}{*}{Location of Object} 
& \textbf{w Gaze (Distractors)} 
& \textbf{w/o Gaze (Distractors)} \\
\cmidrule(lr){2-3}
 & \textcolor{gray}{30k} & \textcolor{gray}{30k} \\
\midrule
Between plate and ramekin    & 96.7 & 78.3 \\
Next to ramekin              & 95.0 & 80.0 \\
Table center                 & 97.0 & 96.7 \\
On cookie box                & 90.0 & 96.7 \\
In cabinet drawer            & 70.0 & 70.0 \\
On ramekin                   & 96.7 & 95.0 \\
Next to cookie box           & 96.7 & 95.0 \\
On stove                     & 83.3 & 76.7 \\
Next to plate                & 85.0 & 42.0 \\
On wooden cabinet            & 93.3 & 60.0 \\
\midrule
\textbf{Overall Avg.}        
& \textbf{89.9} 
& 79.1 \\
\bottomrule
\end{tabular}
}
\vspace{-2mm}
\end{table}

\subsection{Foveated Vision during Training}
\label{sec:foveated}
Prior work has explored using gaze not only as supervision but also to reshape the visual input via foveated rendering, where regions near the gaze location are preserved at high resolution and the periphery is downsampled or blurred~\citep{kim2024multitaskrealrobotdatagaze,chuang2025lookfocusactefficient,li2024virt}. Following this idea, we implement a simple variant in which, for each timestep and view, we construct a foveated RGB image centered on the peak of the gaze distribution and feed this foveated image directly into the standard visual encoder, without changing any other part of the VLA pipeline.

Under a moderate foveation setting, this variant achieves an overall success rate of \textbf{78.5\%} on LIBERO-Spatial, which is roughly \textbf{8 \% lower} than our original non-foveated baseline (85.9\%). We hypothesize that, in our multi-view manipulation setting, aggressively reducing peripheral detail removes useful contextual cues (e.g., table geometry, supporting surfaces, or alternative grasps) that the policy relies on for precise spatial reasoning. 

\begin{table}[!t]
\centering
\caption{
Per-task success rates on LIBERO Spatial~\citep{liu2023libero} at 30k training steps.
We compare the baseline model, our gaze-regularized model, and a foveated-vision variant.}
\label{tab:libero_spatial_30k_threeway}
\resizebox{\columnwidth}{!}{
\begin{tabular}{lccc}
\toprule
\multirow{2}{*}{Location of Object} 
& \textbf{w/o Gaze} 
& \textbf{w Gaze} 
& \textbf{Foveated} \\
\cmidrule(lr){2-4}
 & \textcolor{gray}{30k} & \textcolor{gray}{30k} & \textcolor{gray}{30k} \\
\midrule
Between plate and ramekin & 83.3 & 100   & 80.0 \\
Next to ramekin           & 85.7 & 100   & 81.3 \\
Table center              & 100  & 100   & 95.7 \\
On cookie box             & 100  & 91.3  & 90.0 \\
In cabinet drawer         & 80   & 73.3  & 65.3 \\
On ramekin                & 100  & 100   & 90.0 \\
Next to cookie box        & 100  & 100   & 94.0 \\
On stove                  & 90   & 90    & 80.7 \\
Next to plate             & 50   & 100   & 44.7 \\
On wooden cabinet         & 70.3 & 100   & 63.3 \\
\midrule
\textbf{Overall Avg.}     & 85.9 & \textbf{95.5} & 78.5 \\
\bottomrule
\end{tabular}
}
\vspace{-2mm}
\end{table}

\subsection{Cross-Viewpoint Robustness}
\label{sec:robustness_cross}





Real-world manipulation often involves partial occlusions or temporary sensor failures. To evaluate robustness under such conditions, we remove one camera view at inference time by replacing its RGB frame with a blank image and measure performance on LIBERO-Spatial. Since models are never trained on missing views, this tests their ability to rely on the remaining cameras and maintain spatial consistency and thus, this scenario evaluates its inherent ability to compensate for missing perceptual input by relying on the remaining views and previously learned cross-view spatial consistency.. Both models experience a performance drop, but the gaze-regularized model consistently retains a higher success rate, indicating that gaze supervision encourages more stable and viewpoint-consistent attention, as shown in Table~\ref{tab:missing_views_30k}.

\begin{figure*}[!t]
  \centering
   \includegraphics[width=\linewidth]{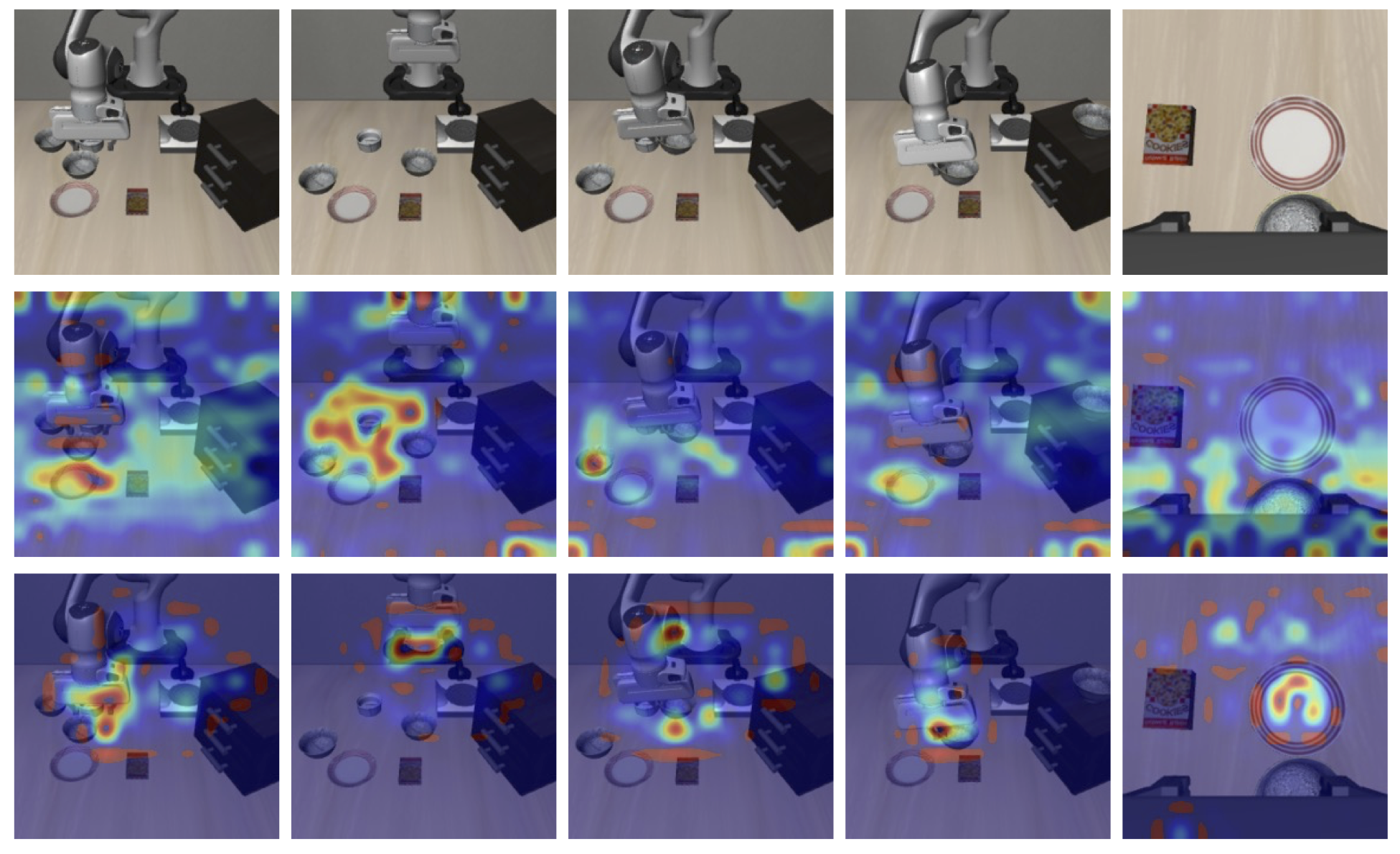}
   \vspace{-3mm}
\caption{\textbf{ Additional Visualisations of Attention.} 
Given the input observation (first), we show the spatial attention from the baseline model (second)  and task-relevant attention produced by our gaze-regularized model (third).}
\vspace{-2mm}
   \label{fig:attn_vis_2}
\end{figure*}

\subsection{Using Gaze Variants}
\label{sec:Gaze-variants}
We further investigate whether different types of gaze supervision influence robustness by evaluating two additional variants: a model trained with DeepGaze \citep{kümmerer2015deepgazeiboosting} (a single-frame gaze predictor) and a Uniform Gaze model where gaze is evenly distributed across all patches. The DeepGaze variant performs moderately well but still falls short of our method, while the Uniform Gaze model exhibits the largest degradation. These trends align with our attention visualizations and Top-k overlap analysis: structured gaze supervision produces sharper, more task-relevant attention, whereas weak or uninformative priors lead to diffuse and unstable attention, reducing performance across tasks. The results are found in Table~\ref{tab:libero_spatial_30k_full}.

\begin{table}[!t]
\centering
\caption{
Per-task success rates on the Missing Views experiment at 30k training steps.
The gaze-regularized model consistently outperforms the baseline across all spatial configurations.}
\label{tab:missing_views_30k}
\resizebox{\columnwidth}{!}{
\begin{tabular}{lcc}
\toprule
\multirow{2}{*}{Location of Object} & \textbf{w Gaze} & \textbf{w/o Gaze} \\
\cmidrule(lr){2-3}
 & \textcolor{gray}{30k} & \textcolor{gray}{30k} \\
\midrule
Between plate and ramekin    & 90.3 & 81.3 \\
Next to ramekin              & 80.7 & 71.0 \\
Table center                 & 90.7 & 81.7 \\
On cookie box                & 70.7 & 62.0 \\
In cabinet drawer            & 69.3 & 60.7 \\
On ramekin                   & 40.7 & 34.7 \\
Next to cookie box           & 69.7 & 60.7 \\
On stove                     & 21.0 & 17.7 \\
Next to plate                & 39.3 & 32.3 \\
On wooden cabinet            & 70.3 & 61.0 \\
\midrule
\textbf{Overall Avg.}        & \textbf{64.3} & 56.3 \\
\bottomrule
\end{tabular}
}
\vspace{-2mm}
\end{table}

\subsection{Using Real Human Gaze for Fine-tuning GLC for Gaze Prediction}
\label{sec:GLC-human}
To enable human-guided gaze prediction for simulation videos, we conducted a data collection study using a screen-based eye tracker which we borrowed briefly for our study. Prior to collection, participants were briefed on each task instruction, ensuring they understood the objective before watching the corresponding simulation video. Their natural eye movements were recorded as they viewed these videos, providing ground truth gaze data for simulation environments. This collected data was then used to fine-tune the GLC model \citep{lai2022eye}, adapting it from its original training on real-world videos to the domain of simulated robotic demonstrations. The resulting model was subsequently used to generate predicted gaze heatmaps for the LIBERO-Spatial benchmark tasks.

To validate the effectiveness of this approach, we compared the performance of our gaze-regularized policy against a baseline trained without gaze supervision. Across the LIBERO-Spatial tasks, the gaze-regularized model consistently outperformed the baseline, demonstrating that even simulation-derived gaze signals provide meaningful guidance for learning visuomotor policies. This performance gap suggests that human attention patterns encode valuable priors about task-relevant visual features that transfer effectively to policy learning.

Importantly, these results were achieved with a relatively modest dataset of human gaze collected specifically for simulation videos. We hypothesize that performance could be further improved by scaling up data collection efforts—incorporating more participants, more diverse tasks, and more finely calibrated eye tracking equipment. Such large-scale, high-quality human gaze data would enable even better adaptation of gaze prediction models to simulation domains, potentially unlocking further gains for gaze-regularized policies. This points to a promising direction for future work: leveraging human attention at scale as a readily accessible form of supervision for robot learning.

\begin{figure*}[t]
  \centering
\includegraphics[width=0.8\linewidth]{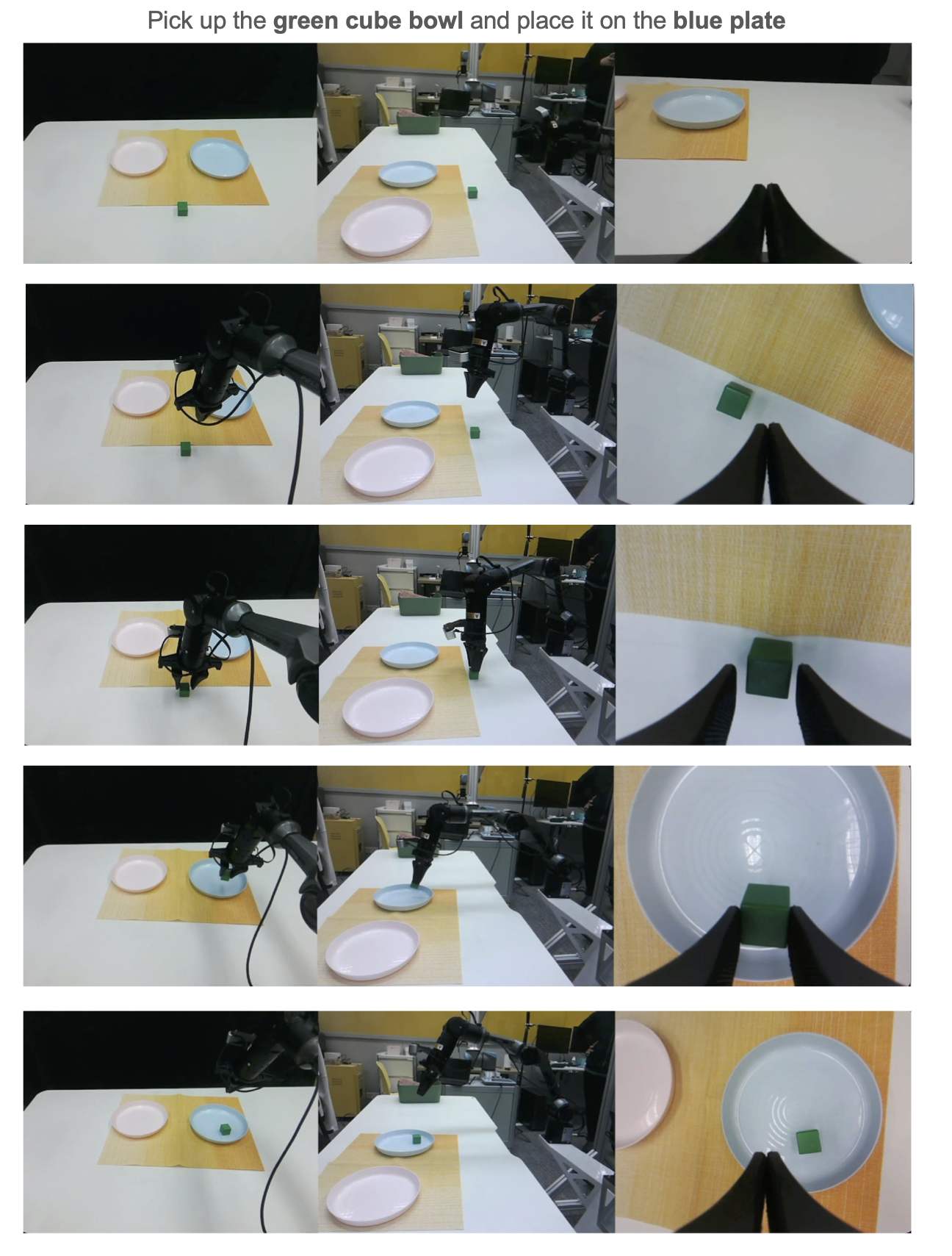}
   \vspace{-2mm}
\caption{\textbf{Visualization of Real-world Task on Aloha Robot} 
In the figure, we provide some frames from a real world task performed using our gaze-regularized policy to show that our method works outside of simulation as well. Here, the task is to pick up the cube and place it on the correct plate.}
\vspace{-2mm}
   \label{fig:results_task_1}
\end{figure*}

\begin{table}[!t]
\centering
\caption{
Per-task success rates on LIBERO Spatial~\citep{liu2023libero} at 30k training steps.
We compare the baseline model, our gaze-regularized model, and the human-gaze-trained variant.}
\label{tab:libero_spatial_30k_human_gaze}
\resizebox{\columnwidth}{!}{
\begin{tabular}{lccc}
\toprule
\multirow{2}{*}{Location of Object} 
& \textbf{w/o Gaze} 
& \textbf{w Gaze} 
& \textbf{Human Gaze} \\
\cmidrule(lr){2-4}
 & \textcolor{gray}{30k} & \textcolor{gray}{30k} & \textcolor{gray}{30k} \\
\midrule
Between plate and ramekin & 83.3 & 100   & 100 \\
Next to ramekin           & 85.7 & 100   & 100 \\
Table center              & 100  & 100   & 100 \\
On cookie box             & 100  & 91.3  & 89.3 \\
In cabinet drawer         & 80   & 73.3  & 78.3 \\
On ramekin                & 100  & 100   & 100 \\
Next to cookie box        & 100  & 100   & 100 \\
On stove                  & 90   & 90    & 90 \\
Next to plate             & 50   & 100   & 100 \\
On wooden cabinet         & 70.3 & 100   & 90 \\
\midrule
\textbf{Overall Avg.}     & 85.9 & \textbf{95.5} & 94.8 \\
\bottomrule
\end{tabular}
}
\vspace{-2mm}
\end{table}

\begin{figure*}[t]
  \centering
\includegraphics[width=0.8\linewidth]{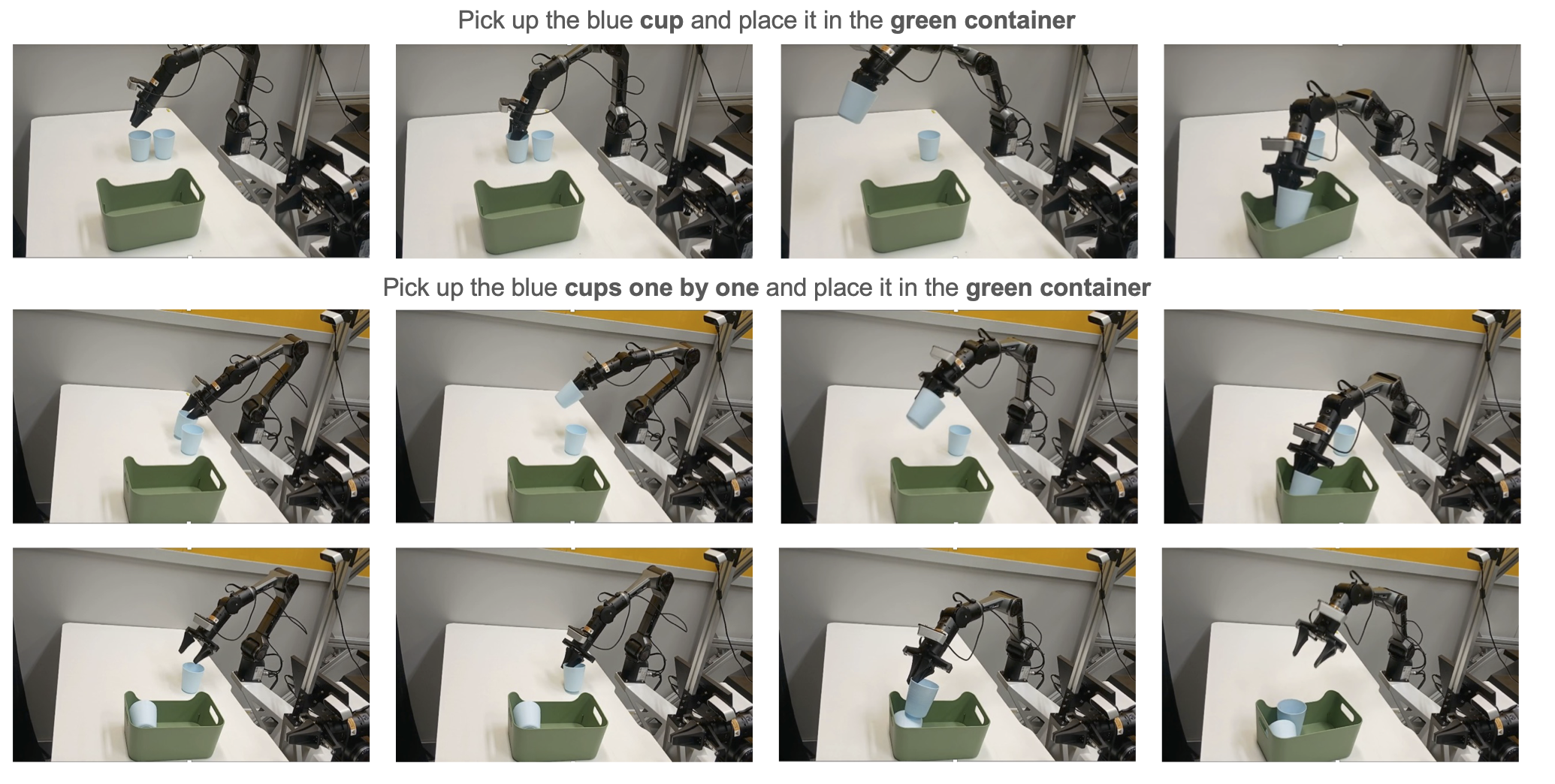}
   \vspace{-2mm}
\caption{\textbf{Visualization of Real-world Task on Aloha Robot} 
In this figure, we present a short horizon task of picking up a cup and placing it in a container(top) and also another longer horizon task to pick up multiple cups one-by-one, and place them in the container. Both visualisations are obtained using our gaze-regularized policy, highlighting its working functionality even in real-world scenarios}
\vspace{-2mm}
   \label{fig:results_task_2}
\end{figure*}

\begin{figure*}[t]
  \centering
\includegraphics[width=0.8\linewidth]{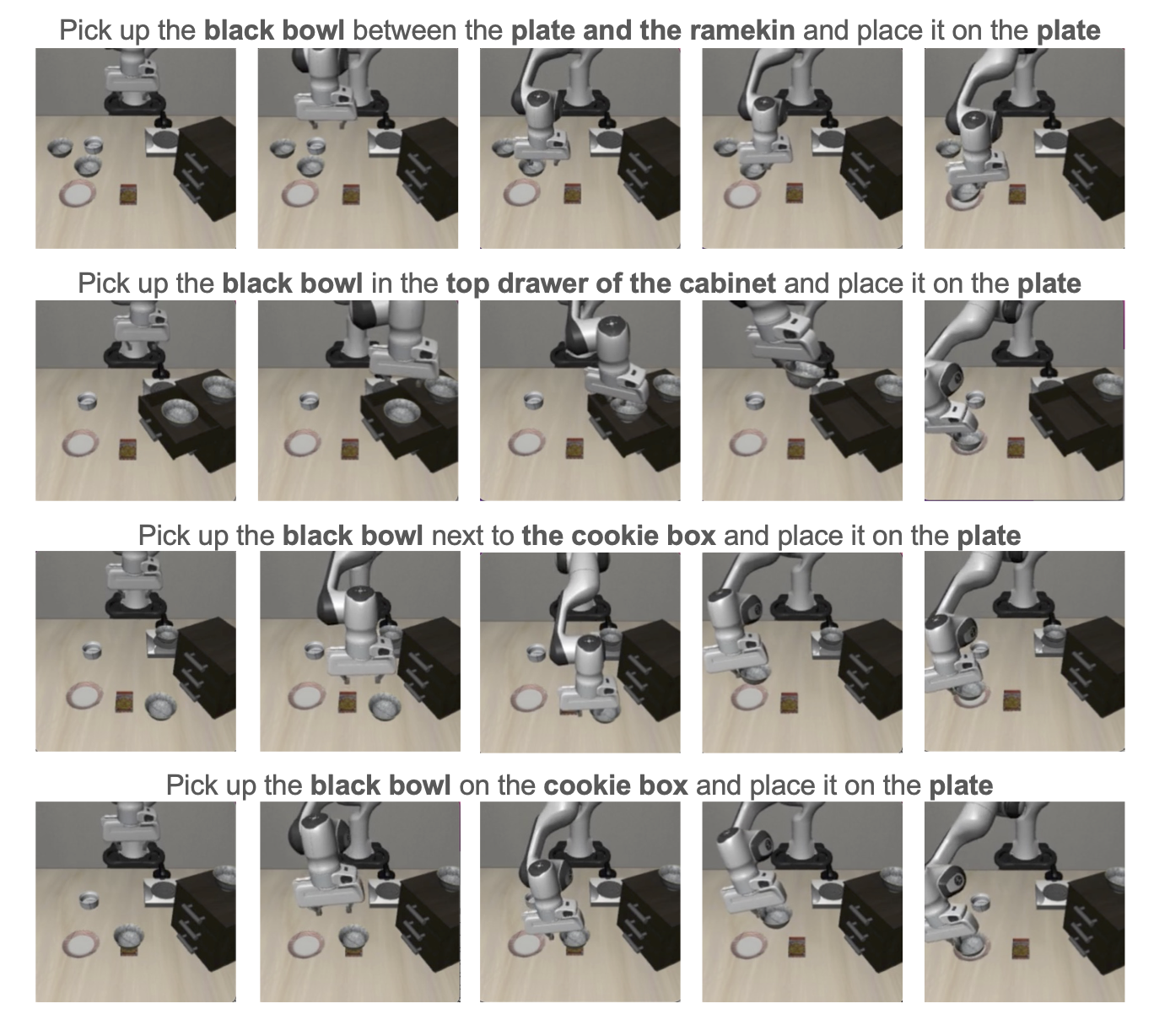}
   \vspace{-2mm}
\caption{\textbf{Visualization Results} 
In the figure, we provide some visualization results to show how the policy performs on the Libero-Spatial \citep{liu2023libero} task suites. We provide the task instructions, and some important frames to show the task success. The baseline model performs admirably, but our method enhances the results by using gaze-regularization.}
\vspace{-2mm}
   \label{fig:results_vis}
\end{figure*}

\section{Pseudocode and Reproducibility}

To facilitate reproduction and adaptation of our method, this appendix summarizes the key implementation components of the gaze-regularized training pipeline. We provide pseudocode for the heatmap-to-token projection used to align gaze with visual tokens to obtain the gaze-prior distribution, and for the overall training loop that integrates gaze regularization into standard VLA optimization.

\subsection{Heatmap-to-Token Projection Pseudocode}

In this section, we provide pseudocode for converting gaze heatmaps produced by the gaze prediction model into patch-level token distributions that are aligned with the transformer’s visual tokens. This procedure is shared across Pi-0 and OpenVLA-based experiments, and can be implemented efficiently using standard tensor operations.


\begin{algorithm}[h]
\caption{Heatmap-to-Token Projection}
\label{alg:heatmap_to_tokens}
\DontPrintSemicolon
\KwIn{Gaze heatmap $H \in \mathbb{R}^{H_g \times W_g}$, patch grid size $P$ (so $N_v = P^2$).}
\KwOut{Patch-level gaze distribution $G \in \mathbb{R}^{N_v}$.}

\BlankLine
\textbf{Step 1: Normalize raw heatmap values.} \\
Compute the sum of all heatmap values:
\[
Z \leftarrow \sum_{x=1}^{H_g} \sum_{y=1}^{W_g} H(x, y).
\]
If $Z = 0$, set $H(x,y) \leftarrow \frac{1}{H_g W_g}$ for all $(x,y)$ (uniform map). Otherwise, normalize:
\[
H(x, y) \leftarrow \frac{H(x,y)}{Z} \quad \forall x,y.
\]

\BlankLine
\textbf{Step 2: Define patch grid.} \\
Let each patch be of size
\[
h_p = \left\lfloor \frac{H_g}{P} \right\rfloor, 
\quad
w_p = \left\lfloor \frac{W_g}{P} \right\rfloor.
\]
For patch indices $u, v \in \{0, \dots, P-1\}$, the spatial region of patch $(u,v)$ is:
\[
\mathcal{P}_{u,v} = 
\{ 
uh_p \le x < (u+1)h_p,\;
vh_p \le y < (v+1)w_p \}.
\]

\BlankLine
\textbf{Step 3: Aggregate heatmap values per patch.} \\
Initialize $G \in \mathbb{R}^{N_v}$ with zeros. \\
\For{$u = 0$ \KwTo $P-1$}{
  \For{$v = 0$ \KwTo $P-1$}{
    $j \leftarrow u \cdot P + v$ \tcp*{flattened patch index}
    \[
    G_j \leftarrow \sum_{(x,y) \in \mathcal{P}_{u,v}} H(x,y).
    \]
  }
}

\BlankLine
\textbf{Step 4: Re-normalize to ensure a valid distribution.} \\
Compute $Z_G \leftarrow \sum_{j=1}^{N_v} G_j$. \\
If $Z_G = 0$, set $G_j \leftarrow \frac{1}{N_v}$ for all $j$. Otherwise:
\[
G_j \leftarrow \frac{G_j}{Z_G} \quad \forall j.
\]

\BlankLine
\textbf{Return} $G$.
\end{algorithm}

\subsection{Training Loop with Gaze Regularization}

We now provide pseudocode for the full training loop, including: (i) multimodal data loading, (ii) synthetic gaze generation via the GLC network, (iii) heatmap-to-token projection, and (iv) optimization with the combined action and gaze-regularization losses. The procedure is shared across all experiments (Pi-0 and OpenVLA backbones), with minor architecture-specific details encapsulated inside the policy forward pass.

\begin{algorithm}[h]
\caption{Training Loop with Gaze Regularization}
\label{alg:training_loop_gaze}
\DontPrintSemicolon
\KwIn{%
    Policy $\pi_\theta$ (VLA model), \\
    Gaze prediction model $\phi_{\text{gaze}}$, \\
    Dataset $\mathcal{D}$ of episodes $\{(I_{1:n,t}, \ell_t, q_t, A_t^\ast)\}$, \\
    Temporal window size $T$ for gaze aggregation, \\
    Regularization scale $\lambda$, \\
}
\KwOut{Trained parameters $\theta^\ast$.}

\BlankLine
\textbf{Initialize} model parameters $\theta$ and optimizer state. \\
\textbf{Repeat} for each training step:

\begin{enumerate}[leftmargin=1.5em]
    \item Sample a batch of timesteps and episodes from $\mathcal{D}$:
    \[
        \{(I_{1:n,t}, \ell_t, q_t, A_t^\ast)\}_{b=1}^B.
    \]

    \item \textbf{Compute synthetic gaze heatmaps.} \\
    For each view $i \in \{1,\dots,n\}$ and each example in the batch, construct a temporal window of frames:
    \[
        \{ I_{i, t-T}, \dots, I_{i,t}, \dots, I_{i, t+T} \}.
    \]
    Pass this sequence through the GLC gaze model:
    \[
        [H_{i,t-T}, \dots, H_{i,t}] 
        \leftarrow 
        \phi_{\text{gaze}}(\{ I_{i, t-T}, \dots, I_{i,t} \}).
    \]

    \item \textbf{Temporal aggregation of gaze.} \\
    Aggregate the per-frame heatmaps around time $t$ using a weighted average:
    \[
        \tilde{H}_{i,t} = \sum_{\delta=-T}^{T} w_\delta H_{i,t+\delta}, 
        \quad
        \sum_{\delta=-T}^{T} w_\delta = 1.
    \]
    This yields a temporally smoothed gaze heatmap per view and frame.

    \item Convert the aggregated heatmap $\tilde{H}_{i,t}$ into a patch-level distribution ($G_{i,t}$)
    
    \item 
    Feed the multimodal observation into the VLA model:
    \[
        A_t = \pi_\theta(I_{1:n,t}, \ell_t, q_t),
    \]
    obtaining predicted action sequences $A_t$. 
    \[
        S_t = \{ S_{i,t} \}_{i=1}^n,
    \]
    where $S_{i,t} \in \mathbb{R}^{N_v}$ is the spatial attention over visual tokens for view $i$.


    \item For each batch element and each view, compute the KL divergence between the gaze prior and the model attention.
\end{enumerate}

\BlankLine
\textbf{Until} convergence or maximum training steps. \\
\textbf{Return} $\theta^\ast$.
\end{algorithm}

\paragraph{Inference.}
At test time, we discard the entire gaze branch:
no gaze model is invoked and no gaze distributions are computed. The policy operates as:
\[
A_t = \pi_{\theta^\ast}(I_{1:n,t}, \ell_t, q_t),
\]
relying only on visual, language, and proprioceptive inputs. The effect of gaze supervision is fully encoded in $\theta^\ast$, manifesting as gaze-aligned internal attention without any inference-time overhead.

\begin{figure*}[t]
  \centering
\includegraphics[width=0.8\linewidth]{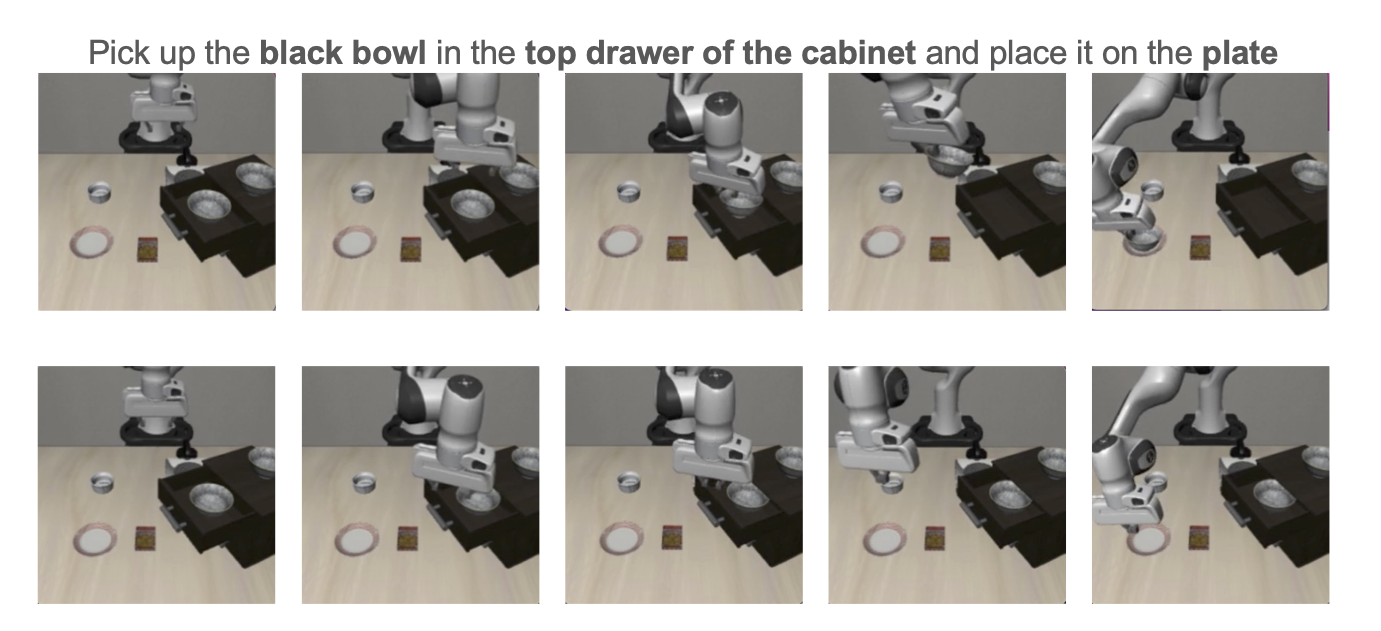}
   \vspace{-2mm}
\caption{\textbf{Failure Case.} We show a failure example from the Libero-Spatial \citep{liu2023libero} task suite. In this task, the baseline model outperforms the gaze-regularized model, suggesting that stronger or more accurate gaze priors could further improve reliability. The bottom sequence illustrates the failure case where the robot hand fails to grab the bowl in the top drawer and proceeds to carry out the intended action.}
\vspace{-2mm}
   \label{fig:results_fail}
\end{figure*}

\section{Summary of Additions}

This supplementary document provides a set of analyses and implementation details that deepen and broaden the claims made in the main paper. We briefly summarize the key additions below and how they support our core hypotheses, and conclude with a discussion of our work.

\paragraph{Clarified notation and methodological details.}
We introduce a consolidated symbol table (Table~\ref{tab:notation_summary}) and expanded descriptions of how visual tokens, language tokens, and gaze-derived distributions interact within the VLA architecture. In particular, we detail how final-layer vision--language cross-attention is extracted, how it relates to action prediction, and why this layer is the most semantically meaningful target for gaze regularization.

\paragraph{Quantitative and qualitative evidence of attention--gaze alignment.}
Beyond task success rates, we define a Top-$k$ attention--gaze overlap metric that directly measures how well the model’s internal attention aligns with gaze-derived priors. Additional visualization of attention maps further illustrate that gaze regularization produces sharper, more task-relevant, and anticipatory attention patterns which aids the action prediction process.

\paragraph{Analysis of synthetic gaze quality.}
We discuss the properties and reliability of the synthetic gaze used in our experiments, motivated by the constraints of existing robotic datasets. Comparisons against alternative gaze priors (e.g., uniform distributions or weaker gaze models) show that performance gains are tied to the \emph{structure} and \emph{quality} of the gaze signal, rather than to generic regularization alone.

\paragraph{Generalization and robustness experiments.}
We extend the evaluation to settings that more closely resemble real-world deployment: (i) linguistic perturbations that alter the phrasing of task instructions, and (ii) cross-viewpoint degradation where one camera input is removed. These experiments demonstrate that gaze-regularized models maintain stronger performance under both language and viewpoint perturbations, highlighting improved robustness and cross-view spatial coherence.

\paragraph{Reproducibility and implementation transparency.}
Finally, we provide pseudocode for the heatmap-to-token projection and for the full training loop with gaze regularization, along with additional implementation notes. These details are intended to make it straightforward to reproduce our results and to adapt the proposed regularization strategy to other VLA architectures and datasets.

Together, these additions reinforce the central message of the our work that incorporating gaze-derived supervisory signals and human priors into VLA training not only improves task performance under standard conditions but also leads to more interpretable, better grounded, and more robust robotic manipulation policies.

\begin{figure}[h]
\vspace{-2mm}
  \centering
   \includegraphics[width=\linewidth]{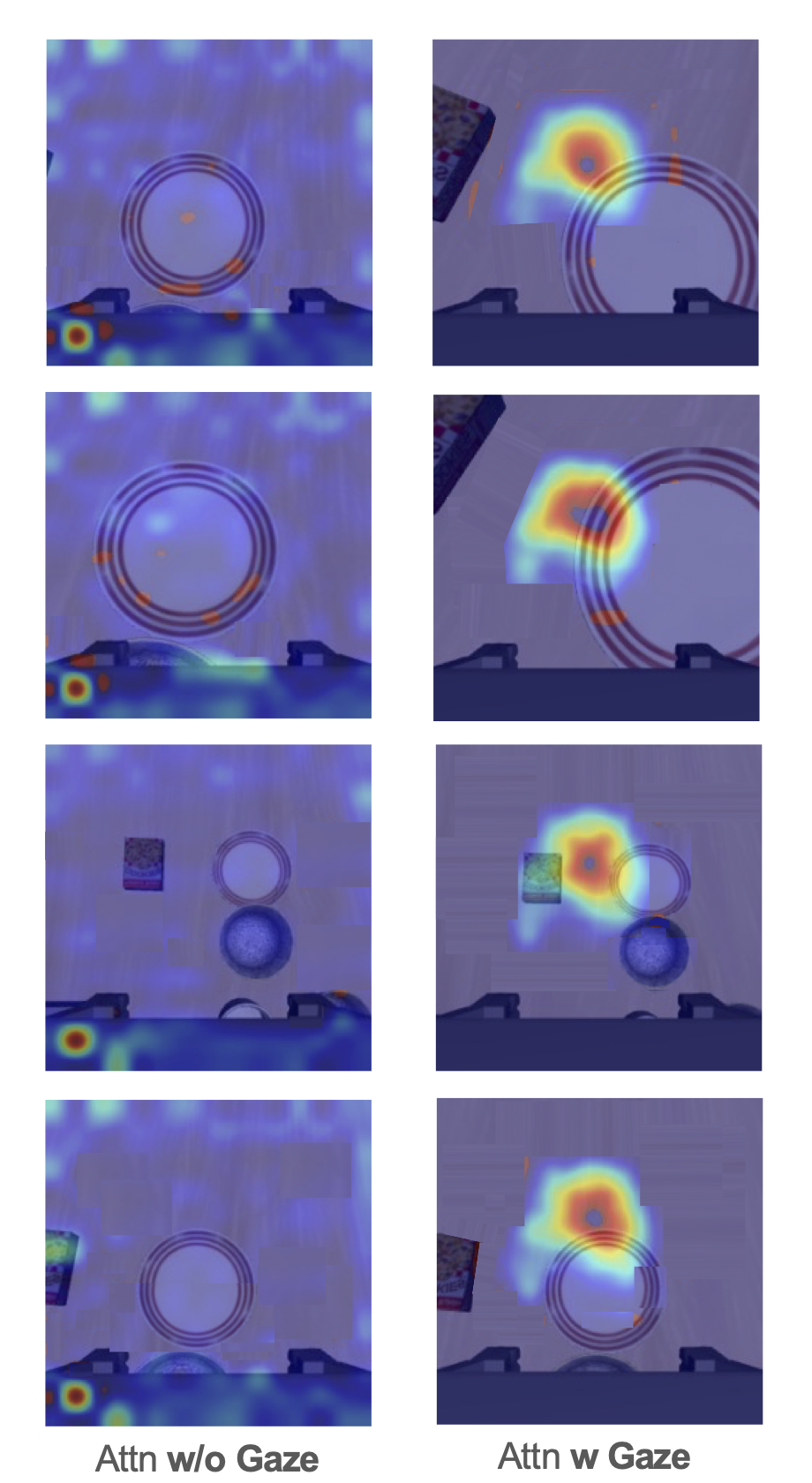}
   \vspace{-2mm}
\caption{\textbf{Attention Comparison.} The baseline model displays diffuse attention spread across the scene, with a single sharp point that is largely task-irrelevant. In contrast, the gaze-regularized model produces noticeably sharper, more concentrated, and consistently task-relevant attention, leading to clearer visual grounding for the instructed action.}
\vspace{-2mm}
   \label{fig:aggregation_new}
\end{figure}

\begin{figure}[!t]
    \centering
    \includegraphics[width=1.0\linewidth]{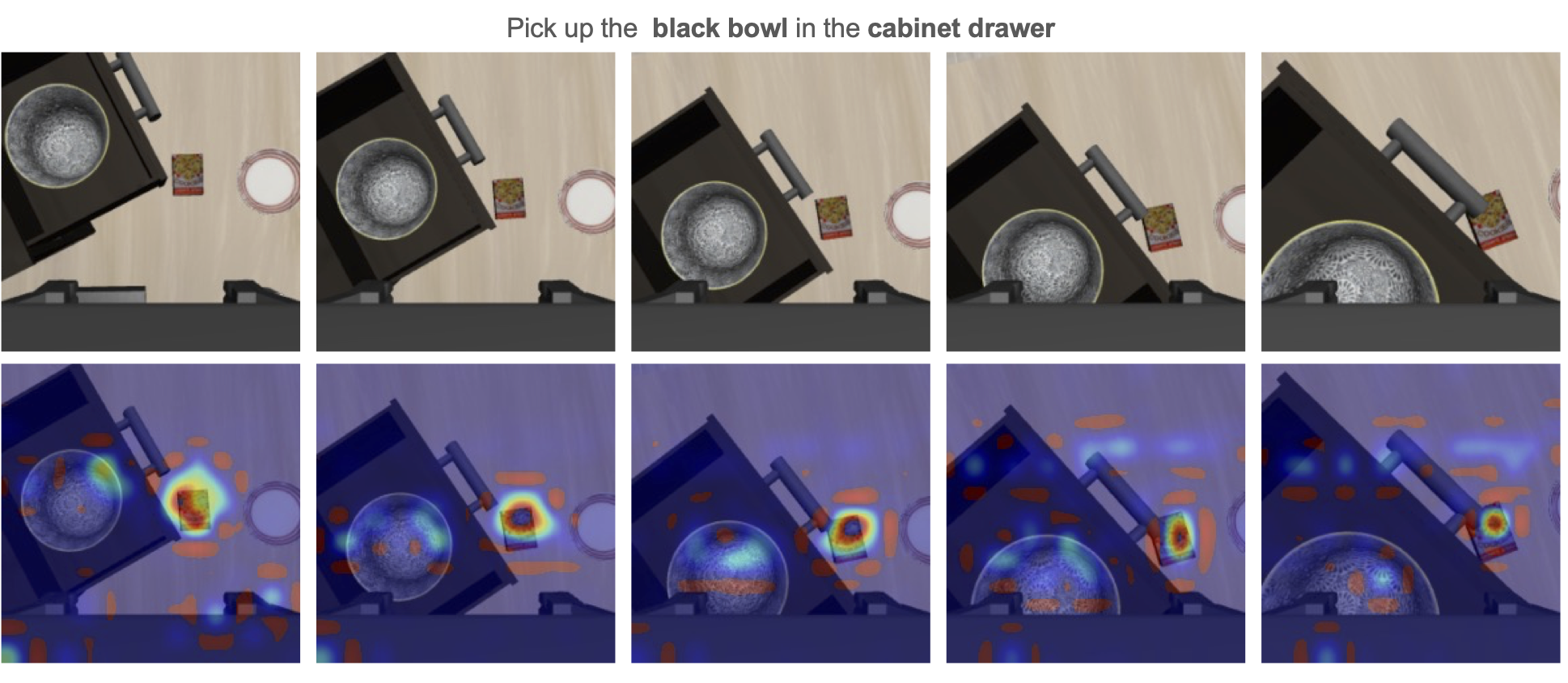}
    \vspace{-5mm}
    \caption{
    \textbf{Visualisation during a failure case.} In this figure, we provide a visualisation of attention during a specific case of failure, where it can be seen that even though the task is to pick up the bowl, attention is not properly distributed on the bowl but rather than on the cabinet handle. Such cases can be mitigated using a better predictor or using a model trained with human supervision on simulated videos 
    }
    \vspace{-3mm}
    \label{fig:teaser}
\end{figure}

\paragraph{Discussion and Limitations}
Our work presents a simple, modular, and architecture-agnostic strategy for improving action prediction in VLA models by incorporating a human-inspired gaze prior during training. The method requires no modification to the underlying VLA design and can be integrated as a lightweight regularization term, making it immediately applicable to a wide range of existing architectures. By guiding the model’s spatial attention toward task-relevant regions-mirroring how humans fixate during manipulation-the policy develops more structured visual grounding, sharper and more discriminative attention maps, and ultimately more reliable action prediction. Across a comprehensive set of experiments, we observe consistent improvements over the baseline model, including enhanced robustness under perturbations, degraded viewpoints, and alternative evaluation protocols. These results highlight that gaze provides a compact yet powerful supervisory signal for spatial reasoning in multimodal transformers. Furthermore, our quantitative and qualitative analyses demonstrate a clear link between sharper attention distributions and improved downstream task success, reinforcing the interpretability of our approach.\\

While promising, our method also opens several avenues for future refinement. First, the synthetic gaze model used in our experiments-though effective-remains an approximation of real human fixation behavior. A more advanced predictor, or one trained directly on teleoperated demonstrations with ground-truth eye-tracking, could further elevate the quality and temporal precision of gaze heatmaps, strengthening the supervisory signal.  Second, our framework currently focuses on RGB-based multi-view perception; extending gaze regularization to richer modalities such as depth, point clouds, or tactile signals may offer additional benefits, particularly in tasks with complex geometry or occlusions. Third, although our approach is inference-free and directly compatible with real-world deployment, we have not yet evaluated it on a physical robot. A hardware implementation would provide valuable insight into how gaze-aligned attention behaves under real-world variations, including lighting changes, hand occlusions, and workspace clutter. Finally, the interaction between gaze priors and large-scale pretraining remains an open question: future work could explore how gaze can be integrated into foundation-model pretraining pipelines or combined with other forms of human supervision, such as demonstrations or language rationales.\\

Overall, our findings illustrate that gaze offers a powerful, interpretable, and low-cost source of inductive bias for VLA training. While there is room for further improvement-especially in gaze quality, multimodal integration, and real-world evaluation-our framework represents a meaningful step toward more perceptually grounded, human-aligned, and robust robotic manipulation policies.

\section{LLM Usage}
We acknowledge the use of LLM in our work for sentence-level re-writing occasionally in our paper to improve the readability, and for suggestions about synonyms, word usage and how to structure and arrange the sections and to check for any spelling/typing mistakes. This was done using ChatGPT and DeepSeek.

\clearpage

\end{document}